\documentclass{article} 
\usepackage{nips14submit_e,times}
\usepackage{hyperref}
\usepackage[lmargin=1.25in,rmargin=1.25in,tmargin=1.1in,bmargin=1.1in]{geometry}
\usepackage{url}
\usepackage{graphicx,mathrsfs}
\usepackage{amsmath}
\usepackage{amssymb}
\usepackage{amsfonts}
\usepackage{cite}
\usepackage{paralist}
\usepackage{verbatim}
\usepackage{caption}
\usepackage{subcaption}
\usepackage{stackrel}
\usepackage{etoolbox} 
\usepackage{sidecap}
\usepackage{enumitem}
\usepackage[compact]{titlesec}
\usepackage{bibunits}
\defaultbibliography{../bibtex/bibtex}
\defaultbibliographystyle{abbrv}

\newcommand*{\eqnameformat}[1]{%
  \textsf{#1}%
}

\makeatletter
\@ifdefinable{\org@maketag@@@}{%
  \let\org@maketag@@@\maketag@@@
  \renewcommand*{\maketag@@@}[1]{%
    \org@maketag@@@{%
      \@ifundefined{eq@name}{#1}{%
        \begin{tabular}[t]{@{}r@{}}%
          #1\tabularnewline
          \eqnameformat{\@nameuse{eq@name}}%
        \end{tabular}%
      }%
    }%
  }%
}
\newif\ifeqname@star
\newcommand*{\eqname}{%
  \@ifstar{\eqname@startrue\eqname@}{\eqname@starfalse\eqname@}%
}
\newcommand*{\eqname@}[2][]{%
  \gdef\eq@name{#1}%
  \ifx\eq@name\@empty
  \else
    \begingroup
      \@ifundefined{GetTitleString}{%
        \gdef\@currenteqlabelname{#2}%
      }{%
        \GetTitleString{#2}%
        \global\let\@currenteqlabelname\GetTitleStringResult
      }%
      \let\@currentlabelname\@currenteqlabelname
      \label{#1}%
    \endgroup
  \fi
  \gdef\eq@name{#2}%
  \ifx\eq@name\@empty
    \global\let\eq@name\relax
  \else
    \ifeqname@star
      \gdef\eq@name{\llap{#2}}%
    \fi
  \fi
}
\@ifdefinable{\org@make@display@tag}{%
  \let\org@make@display@tag\make@display@tag
  \def\make@display@tag{%
    \@ifundefined{@currenteqlabelname}{}{%
      \let\@currentlabelname\@currenteqlabelname
    }%
    \org@make@display@tag
  }%
}
\let\eq@name\relax
\let\@currenteqlabelname\relax
\g@addto@macro\displ@y@{%
  \global\let\eq@name\relax
  \global\let\@currenteqlabelname\relax
}
\@ifdefinable{\org@math@cr@@}{%
  \let\org@math@cr@@\math@cr@@
  \def\math@cr@@[#1]{%
    \org@math@cr@@[{#1}]%
    \noalign{%
      \global\let\eq@name\relax
    }%
  }%
}
\@ifdefinable{\org@eqref}{%
  \let\org@eqref\eqref
  \renewcommand*{\eqref}[1]{%
    \begingroup
      \let\eq@name\relax
      \org@eqref{#1}%
    \endgroup
  }%
}
\g@addto@macro\equation{%
  \eqname{}%
}
\makeatother

\newtheorem{theorem}{Theorem}

\newtheorem{lemma}[theorem]{Lemma}

\newcommand{\beq}{\begin{equation}}
\newcommand{\eeq}{\end{equation}}
\newcommand{\bea}{\begin{eqnarray}}
\newcommand{\eea}{\end{eqnarray}}
\newcommand{\bean}{\begin{eqnarray*}}
\newcommand{\eean}{\end{eqnarray*}}
\newcommand{\bit}{\begin{itemize}}
\newcommand{\eit}{\end{itemize}}
\newcommand{\ben}{\begin{enumerate}}
\newcommand{\een}{\end{enumerate}}
\newcommand{\blem}{\begin{lemma}}
\newcommand{\elem}{\end{lemma}}
\newcommand{\bthm}{\begin{thm}}
\newcommand{\ethm}{\end{thm}}
\newcommand{\bpf}{\begin{proof}}
\newcommand{\epf}{\end{proof}}
\newcommand{\qed}{\hfill$\blacksquare$}

\title{When is it Better to Compare than to Score?}
\author{Nihar B. Shah \qquad Sivaraman Balakrishnan \qquad~~ Joseph Bradley~~ \\ \textbf{Abhay Parekh \qquad ~~Kannan Ramchandran \qquad Martin Wainwright}\\University of California, Berkeley}

\makeatletter
\newenvironment{subtheorem}[1]{%
  \def\subtheoremcounter{#1}%
  \refstepcounter{#1}%
  \protected@edef\theparentnumber{\csname the#1\endcsname}%
  \setcounter{parentnumber}{\value{#1}}%
  \setcounter{#1}{0}%
  \expandafter\def\csname the#1\endcsname{\theparentnumber.\Alph{#1}}%
  \ignorespaces
}{%
  \setcounter{\subtheoremcounter}{\value{parentnumber}}%
  \ignorespacesafterend
}
\makeatother
\newcounter{parentnumber}

\newcommand{\sign}{\textrm{sign}}

\newcommand{\numstud}{d}
\newcommand{\indicator}[1]{\mathbf{1}\{#1\}}

\newcommand{\numobs}{n}

\newcommand{\wt}{w}

\newcommand{\markord}{^{(o)}}
\newcommand{\markcard}{^{(c)}}

\newcommand{\noisestd}{\sigma}
\newcommand{\noisestdo}{\noisestd_o}
\newcommand{\noisestdc}{\noisestd_c}

\newcommand{\eps}{\epsilon}

\newcommand{\reals}{\mathbb{R}}

\newcommand{\obs}{y}
\newcommand{\packsize}{e^{\beta \numstud}}
\newcommand{\packdmin}{\alpha}
\newcommand{\cumlap}{L} 
\newcommand{\cumlapinv}{K} 
\newcommand{\cumlapnorm}{\hat{\Sigma}}

\newcommand{\cumlapinvnorm}{\cumlapnorm^\dagger}
\newcommand{\diff}{\mathbf{x}} 
\newcommand{\diffdirect}{\mathbf{u}}
\newcommand{\diffmx}{X} 
\newcommand{\Lnormsqr}[3]{(#1-#2)^T #3 (#1-#2)}

\newcommand{\Lnorm}[2]{\|#1\|_{#2}}
\newcommand{\loss}{\ell}
\newcommand{\epsvec}{\boldsymbol{\epsilon}}
\newcommand{\wtopt}{\mathbf{\wt^*}}
\newcommand{\wtest}{\mathbf{\hat{\wt}}}
\newcommand{\minimax}{\mathfrak{M}}
\newcommand{\tstone}{{\sc Thurstone}}
\newcommand{\dir}{{\sc Cardinal}}
\newcommand{\pair}{{\sc Paired Linear}}
\newcommand{\btl}{{\sc BTL}}

\newcommand{\norm}[1]{\lVert#1\rVert_2}
\newcommand{\inprod}[2]{\ensuremath{\langle #1 , \, #2 \rangle}}
\newcommand{\trace}[1]{\mathrm{tr}(#1)}
\newcommand{\kl}[2]{\ensuremath{D_{\mathrm{KL}}(#1\|#2)}}
\newcommand{\mnorm}[2]{\|#1\|_{#2}}
\newcommand{\opnorm}[1]{\|#1\|_{\mathrm{op}}}
\newcommand{\amax}{\operatornamewithlimits{arg~max}}
\newcommand{\amin}{\operatornamewithlimits{arg~min}}

\newcommand{\cdf}{\Phi}
\newcommand{\pdf}{\phi}
\newcommand{\wmax}{B}
\newcommand{\betacard}{0.13}
\newcommand{\alphacard}{0.15}

\newcommand{\const}{c}

\nipsfinalcopy 

\begin{document}
\begin{bibunit}
\maketitle
\vspace{-.1cm}
\begin{abstract}
When eliciting judgements from humans for an unknown quantity, one often has the choice of making direct-scoring (\emph{cardinal}) or comparative (\emph{ordinal}) measurements. 
In this paper we study the relative merits of either choice, providing empirical
and theoretical guidelines for the selection of a measurement scheme.
We provide empirical evidence based on experiments on Amazon
Mechanical Turk that in a variety of tasks, (pairwise-comparative) ordinal measurements
have lower per sample noise and are typically faster to elicit 
than cardinal ones. Ordinal measurements however
typically provide less information.
We then consider the popular Thurstone and Bradley-Terry-Luce (BTL) models for ordinal measurements
and characterize the minimax error rates for estimating the unknown quantity. 
We compare these minimax error rates to those under cardinal measurement 
models and quantify for what noise levels ordinal measurements are better. 
Finally, we revisit the data collected from our experiments and show that fitting these models 
confirms this prediction: for tasks where the noise in
ordinal measurements is sufficiently low, the ordinal approach results in smaller errors in the estimation.
\end{abstract}


\section{Introduction}\label{sec:introduction}

\begin{figure*}[b]
\centering
\begin{subfigure}{.3\textwidth}
\includegraphics[width=\textwidth]{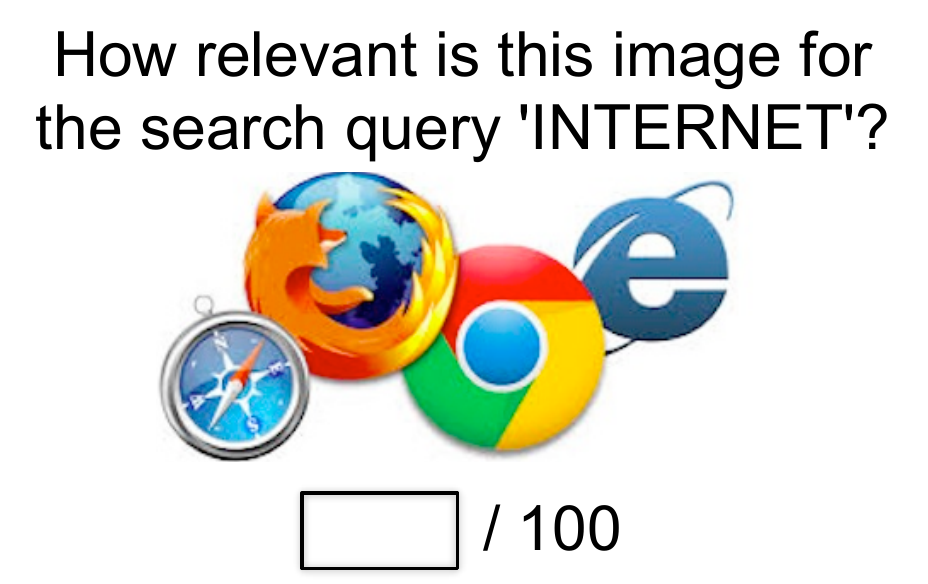}
\caption{Cardinal interface}
\label{fig:searchRelevance_cardinal}
\end{subfigure}
\hspace{.1\textwidth}
\begin{subfigure}{.3\textwidth}
\includegraphics[width=\textwidth]{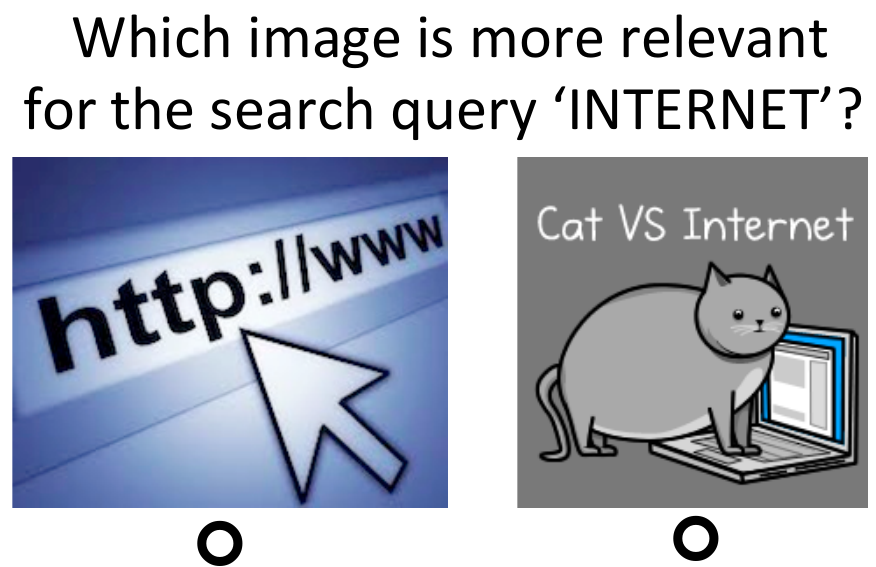}
\caption{Ordinal interface}
\label{fig:searchRelevance_ordinal}
\end{subfigure}
\caption{Examples of cardinal and ordinal interfaces for a task on relevance rating.}
\label{fig:searchRelevance}
\end{figure*}

Eliciting judgements or knowledge about unknown quantities from non-expert humans is commonplace in many domains of society today. This has been facilitated by the emergence of several new `crowdsourcing' platforms such as Amazon Mechanical Turk, that have become powerful, low-cost tools for collecting human knowledge and judgements. However, this low cost comes at the price of noise, due to the unreliability in the crowd response. This paper addresses this issue of noise at the source by studying how responses should be elicited.

We consider a setting in which humans perform evaluations that have numeric answers. Examples include a crowdsourcing task that involves counting the number of malaria parasites in an image of a blood smear~\cite{luengo2012crowdsourcing}, or a peer-grading task that involves students assigning grades to homeworks submitted by other students~\cite{piech2013tuned}. A standard design of such a task takes a \textit{cardinal} approach where the evaluators directly enter numeric scores as the answers. This is illustrated by the example in Figure~\ref{fig:searchRelevance_cardinal} where the subject is asked to rate the relevance of an image for the search query `Internet' as a numeric entry between 0 and 100.

Alternatively, one could take an \textit{ordinal} approach, asking the evaluator to compare (or rank order) multiple items. Such an ordinal method is illustrated in Figure~\ref{fig:searchRelevance_ordinal} where  the evaluator is shown a pair of images, and is asked to select the one that is more relevant for the search query `Internet'. In this paper, we restrict our attention to comparisons of only pairs of items in the ordinal setting.


Cardinal measurements allow for more precise measurements; in Figure~\ref{fig:searchRelevance}, one cardinal measurement can take 100 values, whereas one ordinal measurement provides a single bit. One may be tempted to go even further and argue that ordinal measurements necessarily give less information, for one can always convert a set of cardinal measurements into ordinal, simply by ordering the measurements by value.
The \textit{data processing inequality}~\cite[Section 2.8]{cover2012elements} then suggests that an estimation procedure on any manipulation of the data cannot perform better than estimating from the original data.
This may lead one to conclude that the ordinal data cannot yield superior results.

In contrast, ordinal measurements avoid calibration issues that are frequently encountered in cardinal measurements~\cite{tsukida2011analyze}, such as the evaluators' inherent (and possibly time-varying) biases, or tendencies to give inflated or conservative evaluations.
Ordinal measurements are also recognized to be easier or faster for humans to make~\cite{barnett2003modern,stewart2005absolute}, allowing for more evaluations for the same level of time, effort and perhaps cost as well.

The lack of clarity regarding when to use a cardinal versus an ordinal approach forms the motivation for this paper. We first address the fundamental question of how much information we gain from each type of measurement. In extensive experiments on a variety of tasks, we find that the average per-sample noise is often significantly higher in cardinal measurements than in ordinal ones.
In other words, \emph{the data processing inequality does not apply when comparing cardinal and ordinal work from humans}.

While revealing, this still leaves two questions: Can we still make reliable estimates from paired comparisons? How much lower does the noise have to be for comparative measurements to be preferred over cardinal measurements? To address this, we invoke theoretical models for pairwise and cardinal measurements. We study the Thurstone (Case V) model~\cite{thurstone1927law}, one of the most widely used models in both theory~\cite{bramley2005rank,krabbe2008thurstone} and practice~\cite{swets1973relative,chessbase2007elo,herbrich2007trueskill}. We will show that it is indeed possible to perform estimation using pairwise comparisons, and via \emph{minimax theory} we will quantify the settings in which pairwise comparisons are preferable to cardinal measurements. Minimax theory is a cornerstone of statistical decision theory and is a standard tool
used in the comparison of estimators in a given model. In this paper, we will investigate the utility of this statistical perspective in comparing estimators
\emph{across} cardinal and ordinal models.

We also provide \emph{topology-aware} bounds that incorporate the choice of pairs to be compared for the Thurstone
and other popular pairwise-comparison models. Of particular importance is the popular 
Bradley-Terry-Luce (BTL) model~\cite{bradley1952rank,luce1959individual}. These bounds highlight the influence of the \emph{comparison graph}
on the estimation error.

Finally, we return to the data obtained from our experiments and fit our ordinal and cardinal models. We observe that the estimates produced from the ordinal data are more accurate than those from cardinal data when the ordinal noise is low enough. This suggests the following \textit{practical guideline} in choosing between the cardinal and ordinal methods of data collection, of first estimating the noise in the two approaches by eliciting a few samples where the ground truth is known. The ordinal approach is then preferred if the ordinal noise is ``low enough''. For tasks in which the ordinal approach is preferred, our topology-aware results provide guidelines for the selection of items to compare when given a fixed budget.

%

\section{Experiments Comparing Per-sample Noise in Cardinal and Ordinal}\label{sec:experiments}
It is tempting to argue that a cardinal sample always gives more information than an ordinal sample: given cardinal samples, one can always order them thereby obtaining ordinal values. This argument suggests that an ordinal approach leads to a loss of information, and due to the data-processing inequality, cannot lead to better results. In this section, by means of seven different experiments conducted on Amazon Mechanical Turk (mturk.com), we show that such an argument is flawed. The experiments also provide insights into the per-sample noise in the ordinal and cardinal methods of data collection, which is a metric that the subsequent theory in this paper will also focus on.


Each experiment involved a certain task that was given to 100 human subjects. Each of these subjects was randomly given either the ordinal or the cardinal version of the task. Both versions had the same set of questions, and each question had a numeric answer. In the cardinal version of the task, the subject was required to directly provide this number as the answer. The ordinal version presented the questions in pairs, and for each pair, the subject had to select the one which she believed had a larger number as the answer.

We now describe the tasks presented to the subjects in the seven experiments. The tasks were selected to have broad coverage of several important subjective judgment paradigms such as preference elicitation, knowledge elicitation, audio and visual perception, and skill utilization.

\begin{figure}[t]
\centering
\begin{subfigure}{0.26\textwidth}
\includegraphics[width=\textwidth]{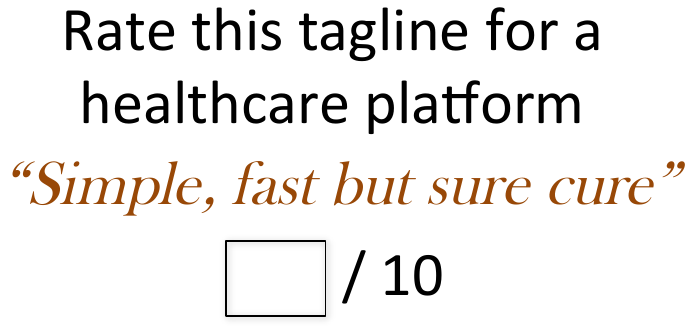}
\caption{}
\label{fig:tagline_cardinal}
\end{subfigure}
\quad
\begin{subfigure}{0.26\textwidth}
\includegraphics[width=\textwidth]{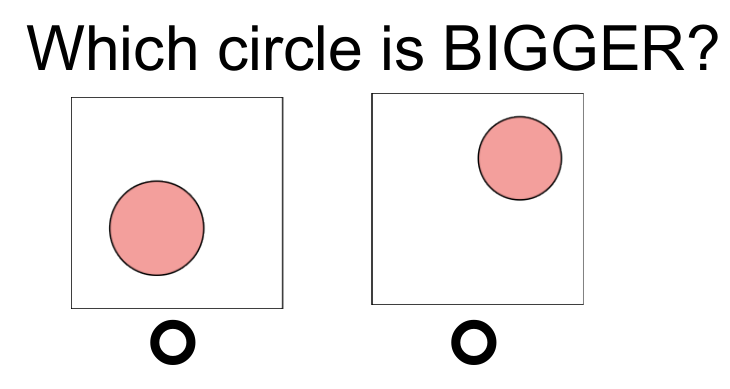}
\caption{}
\label{fig:areaCircle_ordinal}
\end{subfigure}
\quad
\begin{subfigure}{0.27\textwidth}
\vspace{-.1cm}
\includegraphics[width=\textwidth]{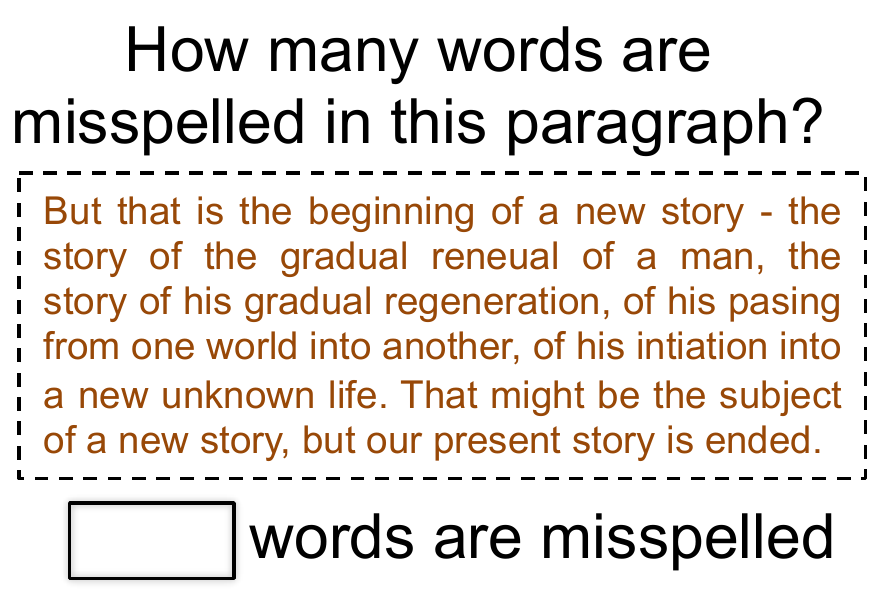}
\caption{}
\label{fig:spelling_cardinal}
\end{subfigure}
\\
\begin{subfigure}{0.24\textwidth}
\includegraphics[width=\textwidth]{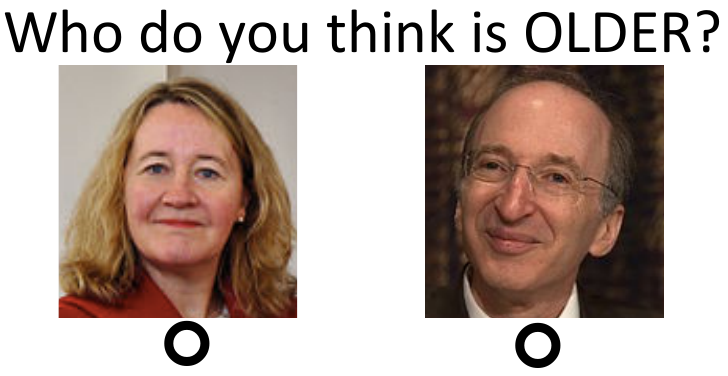}
\caption{}
\label{fig:photoAge_ordinal}
\end{subfigure}
\quad
\begin{subfigure}{0.24\textwidth}
\includegraphics[width=\textwidth]{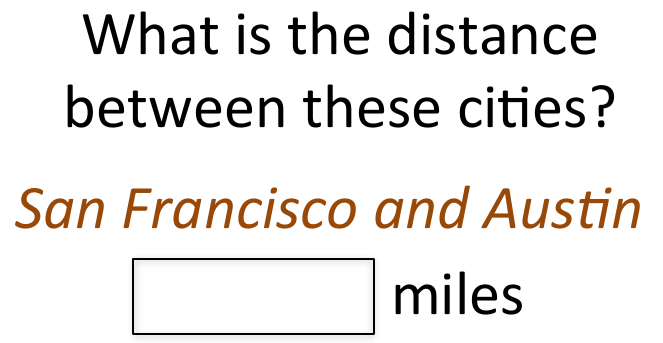}
\caption{}
\label{fig:distanceCities_cardinal}
\end{subfigure}
\quad
\begin{subfigure}{0.2\textwidth}
\includegraphics[width=\textwidth]{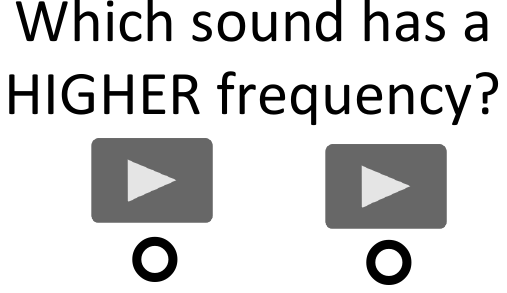}
\caption{}
\label{fig:tones_ordinal}
\end{subfigure}
\caption{Screenshots of the tasks presented to the subjects. For each task, only one version (cardinal or ordinal) is shown here.}
\end{figure}

\begin{table}[b]
\begin{center}
\begin{tabular}{|l|c|c|c|c|c|c|c|}
\hline
Task & Tagline & Circle & Spelling & Age & Distance & Audio & Relevance\\
\hline
Error in Ordinal   & \textbf{29\%} &  \textbf{6\%} & \textbf{40\%} & \textbf{13\%} & \textbf{17\%} & \textbf{20\%} & \textbf{22\%} \\
Error in Cardinal & 31\% & 18\% & 46\% & 17\% & 46\% & 31\% & 27\% \\
\hline
Time in Ordinal & \textbf{251s}  & \textbf{98s}& \textbf{144s} & \textbf{31s}&  \textbf{84s}  & \textbf{66s} & \textbf{105s} \\
Time in Cardinal & 342s& 181s & 525s & 70s& 305s & 134s & 185s\\
\hline
\end{tabular}
\end{center}
\caption{Comparison of the average amount of error when ordinal data was collected directly vs. when cardinal data was collected and converted to ordinal. Also tabulated is the median time (in seconds) taken to complete a task by a subject in either type of task.}
\label{tab:mturk_results}
\end{table}

\textbf{(a) Rating taglines for a product:} A product was described and ten taglines for this product were shown (Figure~\ref{fig:tagline_cardinal}). The subject had to rate each of these taglines in terms of its originality, clarity and relevance to this product.\vspace{2pt}\\
\textbf{(b) Estimating areas of circles:} The task comprised 25 questions. In each question, the subject was shown a circle in a bounding box (Figure~\ref{fig:areaCircle_ordinal}), and the subject was required to identify the fraction of the box's area that the circle occupied.\vspace{2pt}\\
\textbf{(c) Finding spelling mistakes in text:} Eight paragraphs of text were shown, and the subject had to identify the number of words that were misspelled in each paragraph (Figure~\ref{fig:spelling_cardinal}).\vspace{2pt}\\
\textbf{(d) Estimating age of people from photographs:} The subject was shown photographs of ten people (Figure~\ref{fig:photoAge_ordinal}) and was asked to estimate the ages of the ten people.\vspace{2pt}\\
\textbf{(e) Estimating distances between pairs of cities:} The subject was shown sixteen pairs of cities (Figure~\ref{fig:distanceCities_cardinal}) and for each pair, the subject had to estimate the distance between them.\vspace{2pt}\\
\textbf{(f) Identifying sounds:} The subject was presented with ten audio clips, each of which was the sound of a single key on a piano (which corresponds to a single frequency). The subject had to estimate the frequency of the sound in each audio clip (Figure~\ref{fig:tones_ordinal}).\vspace{2pt}\\
\textbf{(g) Rating relevance of the results of a search query:} Twenty results for the query `Internet' for an image search were shown (Figure~\ref{fig:searchRelevance}) and the subject had to rate the relevance of these results with respect to the given query.

Upon obtaining the data from the experiments, we first reduced the cardinal data into ordinal form by comparing answers given by the subjects to consecutive questions. For five of the seven experiments ((b) through (f)), we had access to the ``ground truth'' solutions, using which we computed the fraction of answers that were incorrect in the ordinal and the cardinal-converted-to-ordinal data (any tie in the latter case was counted as half an error). For the two remaining experiments ((a) and (g)) for which there is no ground truth, we computed the `error' as the fraction of (ordinal or cardinal-converted-to-ordinal) answers provided by the subjects that disagreed with each other.

The results are tabulated in Table~\ref{tab:mturk_results} (boldface indicates a better performance). If the data-processing inequality were true, then it would be unlikely for the amount of error in the ordinal setting to be lower than that in the cardinal setting. On the contrary, one can see from Table~\ref{tab:mturk_results} that converting cardinal data to an ordinal form results in a typically higher (and sometimes significantly higher) per-sample error than directly asking for ordinal evaluations. This absence of data-processing inequality may be explained by the argument that the inherent evaluation process in the human subjects is not the same in the cardinal and ordinal cases -- humans do \textit{not} perform an ordinal evaluation by first performing cardinal evaluations and then comparing them (this is why it is often found to be easier to compare than score~\cite{barnett2003modern,stewart2005absolute}). One can also see from Table~\ref{tab:mturk_results} that the amount of time required for cardinal evaluations was typically (much) higher than for ordinal evaluations.

\section{Theoretical Comparison of  Cardinal and Ordinal Measurement Schemes}\label{sec:comparison}
The experiments in the previous section established that the `per-sample noise' in the cardinal setup is typically larger than that in the ordinal setting. However, each ordinal sample, unlike a cardinal value, can provide just one bit of information. This discrepancy is further complicated by the fact that the multitude of samples collected from multiple workers need to be aggregated in order to produce final estimates of the answers. It is thus not clear for a given a problem setting, whether an ordinal or a cardinal method of data collection would yield a superior performance. This section aims at addressing this issue: given that ordinal and cardinal samples have a different nature and amount of noise, which method of data collection will produce a smaller aggregate error?

In this section we focus our attention on the Thurstone (Case V) generative model~\cite{thurstone1927law}, which is one of the most popular models considered in both theory~\cite{bramley2005rank,krabbe2008thurstone,nosofsky1985luce} and practice~\cite{swets1973relative,chessbase2007elo,herbrich2007trueskill}. This model assumes that every item has a certain numeric \emph{quality score}, and a comparison of two items is generated via a comparison of the two qualities in the presence of an additive Gaussian noise.

%
We define a vector $\wtopt \in \mathbb{R}^d$ of qualities, so item
$j \in [\numstud]$ has quality $\wt^*_j$.
Under the Thurstone model we compare pairs of items items. For $i \in [n]$
the outcome
of the $i^{\mathrm{th}}$ 
comparison is $\obs\markord_i \in \{-1,1\}$, where $\obs\markord_i$ is given by
\beq
\label{eqn::thurstone}
\obs\markord_i = \sign(\wtopt^T \diff_i + \eps_i\markord), 
\tag{\tstone}
\eeq
$\eps_i\markord$ is independent Gaussian noise with variance $\noisestd_o^2$,
and $\diff_i \in \mathbb{R}^d$ is a differencing vector
with one entry $+1$, one entry $-1$ and the rest $0$. Observe that the ordinal model
is identifiable only upto a shift in $\wtopt$ so we always assume $\mathbf{1}^T \wtopt = 0.$

The cardinal analogue of this model involves a cardinal evaluation of individual items, where
for $i \in [n]$ the outcome $\obs\markcard_i$ is given by
\beq
\label{eqn::direct}
\obs\markcard_i = \wtopt^T \diffdirect_i + \eps_i\markcard
\tag{\dir}
\eeq
where $\diffdirect_i$ in this case is a coordinate vector with one of its entries equal to $1$ and remaining entries $0$,
and $\eps_i\markcard$  is independent Gaussian noise, with a \emph{different variance} $\noisestd_c^2$.

In order to build intuition on how to compare these models, in this section we focus on a simple scenario.
Subsequently, in Section~\ref{sec::bounds} we consider general settings.
Analogous to the \emph{fixed design regression} setup, we choose the vectors $\diff_i$ a priori.
Suppose that
$n$ is large enough, and that in the ordinal case we compare each pair $n/{d \choose 2}$ times.
In the cardinal case suppose that we evaluate the quality of each item $n/d$ times.
%

To facilitate a comparison between the \dir~and \tstone~models we consider the \emph{minimax risk}. 
In each case a vector $\mathbf{w}$ induces a distribution $\mathbb{P}_{\mathbf{w}}$ from which 
the
observed samples $\{y_1,\ldots,y_n\}$ are drawn (recall that the vectors $\diff_i$ are \emph{fixed}). 
Let 
 $\mathcal{P}$ denote the family of induced distributions and $\mathcal{W}$ denote the set of allowed vectors $\mathbf{w}$.
 An estimator
 $\wtest$ is a (measurable) map from the observed samples to $\mathcal{W}$.
For a semi-norm $\rho$ the minimax risk is 
$$\minimax^\rho_n := \inf_{\wtest} \sup_{\mathbb{P}_{\mathbf{w}} \in \mathcal{P}} \mathbb{E} [\rho(\wtest, \mathbf{w})]$$
where the expectation is taken over the samples $\{y_1,\ldots,y_n\}.$
The minimax risk characterizes the performance of the \emph{best} estimator in the metric induced by $\rho.$
In this section we focus on the case when \mbox{$\rho(\wtest, \mathbf{w}) = \norm{\wtest - \mathbf{w}}^2$}
and we denote the minimax risk as $\minimax^2_n.$ 

With these preliminaries in place we can attempt to ask a basic question for the simple case
of evenly budgeted measurements:
\emph{Given $\numobs$ samples with noise standard-deviation $\noisestdc$ in the cardinal case and $\noisestdo$ in the
ordinal case, is the expected minimax error in the estimation of $\numstud$ items lower in the cardinal case or the ordinal case? }
The following theorem provides an answer for many regimes of $(\noisestdc,\noisestdo)$. 
%
%
\begin{theorem}
\label{thm::cvo}
Suppose that $\numobs$ is large enough and that in the \dir~model we observe each
coordinate $\numobs/\numstud$ times. The minimax risk is
\begin{equation*}
\frac{\minimax^2_n(\textsc{\dir})}{\numstud} = \frac{\numstud \noisestdc^2}{\numobs}.
\end{equation*}
Suppose that $\numobs$ is large enough and that in the \tstone~model we observe
each pair $\numobs/{\numstud \choose 2}$ times. Suppose $\Lnorm{\mathbf{\wt^*}}{\infty}\leq \wmax$, and that $\wmax$ and $\noisestdo$ are known. Let $\cdf$ denote the standard Gaussian c.d.f., and let $\kappa := \cdf(2\wmax/\noisestdo) (1-\cdf(2\wmax/\noisestdo)).$ 
Then the minimax risk is bounded as
\begin{equation*}
0.0008 \kappa \frac{\numstud \noisestdo^2}{\numobs} \leq \frac{\minimax^2_{\numobs}(\textsc{\tstone})}{\numstud}\leq 
\frac{5}{\kappa^2} \frac{\numstud \noisestdo^2}{\numobs}~.
\end{equation*}
\end{theorem}

%
%
%
In the cardinal case when each coordinate is measured the same number of times, the \dir~model reduces to the well-studied normal location model, for which the MLE is known to be the minimax estimator and its risk is straightforward to characterize (see \cite{tpe} for instance). In the ordinal case the result follows from the general treatment in Section~\ref{sec::bounds}. Observe that the \tstone~minimax bounds depend on $\|\wtopt\|_\infty.$ This is related to the strong convexity parameter of the likelihood in the \tstone~model which degrades for increasing $\|\wtopt\|_\infty.$ Informally, this is related to the difficulty of estimating very small (or very large) probabilities that can 
arise in the \tstone~model for large $\|\wtopt\|_\infty.$

\begin{SCfigure}
\centering
\includegraphics[width=.39\textwidth]{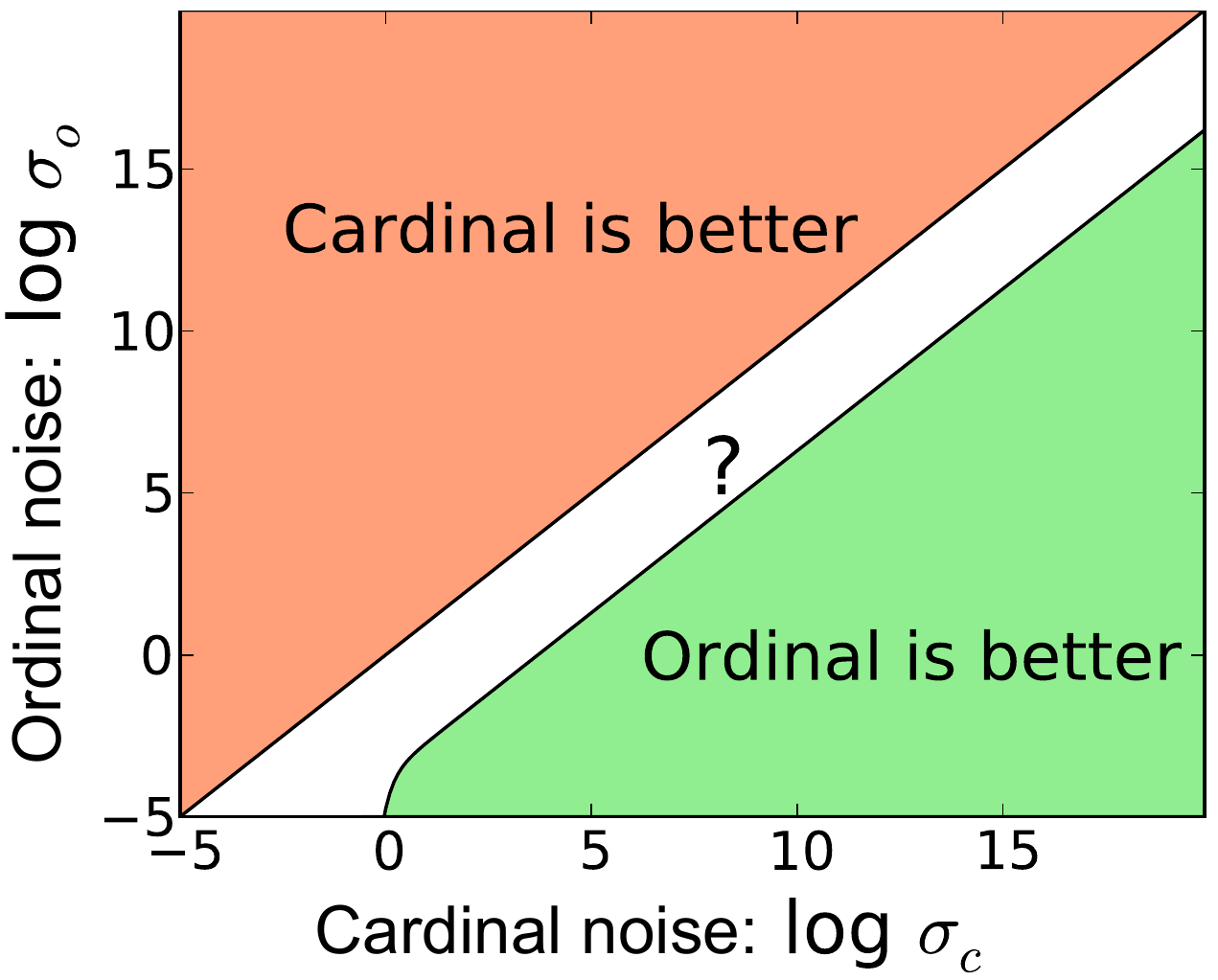}
\caption{Characterizing the regions of $(\noisestdc, \noisestdo)$ where cardinal or ordinal methods lead to a lower minimax error under the \dir~and~\tstone~models respectively.  $\wmax$ is fixed at $1$. The bounds for the \tstone~model are loose when the signal to noise ratio (SNR) is high but relatively tighter at low SNR; the log-log scale of the axes attempts to focus on this low-SNR regime.\vspace{.8cm}}
\label{fig:which_is_better_cardinal_vs_ordinal_estimation}
\end{SCfigure}

Observe from Theorem~\ref{thm::cvo} that the minimax risks in the cardinal and ordinal settings have the same dependency on $\numstud$ and $\numobs$. An ordinal approach of collecting data is thus better overall whenever its per-sample error is ``low enough''. Figure~\ref{fig:which_is_better_cardinal_vs_ordinal_estimation} summarizes the result of Theorem~\ref{thm::cvo}. 

\section{General Bounds and Topology Considerations}
\label{sec::bounds}
In the previous section, we analyzed one paired comparison model, the \tstone~model.  We now provide a more general treatment by considering three models while allowing arbitrary comparisons.
In addition to the \tstone~model
we also provide results for its linear and
logistic analogues:
\beq
\obs_i = \mathbf{\wt^*}^T \diff_i + \eps_i \qquad \textrm{for } i \in [\numobs],\tag{\pair}
\eeq
where $\eps_i$ are i.i.d. $N(0,\sigma^2)$, and 
\beq
\label{eqn::btl}
\mathbb{P}(\obs_i = 1| \diff_i, \mathbf{\wt^*}) = \frac{1 }{ 1 + \exp \left( \frac{- \mathbf{\wt^*}^T \diff_i }{\noisestd}\right) } \qquad \textrm{for } i \in [\numobs]. \tag{\btl}
\eeq
As before, the $\diff_i$'s are difference vectors, and we assume $\mathbf{1}^T \wtopt = 0.$
The second model 
is the popular Bradley-Terry-Luce (\btl) model~\cite{bradley1952rank,luce1959individual}. The \btl~model is also a popular choice for modeling pairwise comparisons~\cite{nosofsky1985luce,atkinson1998asian,koehler1982application,heldsinger2010using,loewen2012testing}, especially since it allows for a computationally simple maximum likelihood inference.  The parameter $\noisestd$ plays the role of a noise parameter, with a higher value of $\noisestd$ leading to more uncertainty in the comparisons. We will assume, under all the models, that the value of $\noisestd$ is known. 
We note in passing that each of these models is a special case of \emph{generalized
linear models} (GLMs) \cite{glmbook}, and that many of the insights here carry over to this general class. 
We defer a detailed treatment of GLMs to an extended version.

In this section we will not assume that items are chosen uniformly at random, rather we provide
bounds in the general case when the measurements are fixed \emph{a priori}. 
This will highlight the central role of the
Laplacian
of the weighted graph of chosen comparisons.
The minimax rate for estimating the underlying quality in $\ell_2^2$ will depend on the
spectral properties of the Laplacian which in turn depends on the \emph{topology} 
of the underlying comparison graph.


In the ordinal models, each measurement is related to a 
\emph{difference} of two quality assessments. 
Observe that the
\emph{covariance matrix} of the measurements is
\bean
\cumlapnorm := \frac{1}{n} \sum_{i=1}^{\numobs} \diff_i \diff_i^T := \frac{\cumlap}{n}
\eean
where $\cumlap$ is
the combinatorial graph Laplacian of the undirected graph
with each edge having a weight equal to the number of times its end points are compared.
We refer to $\cumlapnorm$ as the \emph{standardized} Laplacian.
The standardized Laplacian is positive semi-definite and has at least one zero-eigenvalue corresponding to the 
all ones vector.  We assume 
that the graph induced 
by the comparisons is \emph{connected}, since it is easy to verify that without this the model is not identifiable. 
The covariance matrix induces a semi-norm on vectors in $\mathbb{R}^d$, defined as \mbox{$\mnorm{\mathbf{v}}{\cumlapnorm} = \sqrt{\mathbf{v}^T \cumlapnorm \mathbf{v}}.$} We denote the Moore-Penrose 
pseudo inverse of $\cumlapnorm$ by $\cumlapnorm^\dagger.$
We first focus on the minimax risk of estimating $\wtopt$ in the \emph{squared} semi-norm induced by $\hat{\Sigma}.$
We denote this as $\minimax^{\cumlapnorm}_n$. Theorem~\ref{thm:minimax} below bounds this minimax risk in each of the
three models. To cleanly state our results we make the simplifying assumption that $\numstud > 9$.
\begin{subtheorem}{theorem}\label{thm:minimax}
\begin{theorem}[\pair]\label{thm:pair}
 The minimax rate is bounded as
\begin{equation*}
0.00013 \frac{\numstud \noisestd^2}{\numobs} \leq \minimax^{\cumlapnorm}_n(\textsc{\pair}) \leq 
0.68 \frac{\numstud \noisestd^2}{\numobs}~.
\end{equation*}
\end{theorem}

\begin{theorem}[\tstone]\label{thm:tstone}
Assume that $\Lnorm{\mathbf{\wt^*}}{\infty} \leq \wmax$ (known). Let $\kappa := \cdf(2\wmax/\noisestd) (1-\cdf(2\wmax/\noisestd)),$ and let \mbox{$n \geq \frac{\noisestd^2 \kappa\trace{\cumlapnorm^\dagger}}{0.035 \wmax^2}$.} The minimax rate is bounded as
\begin{equation*}
0.0008 \kappa \frac{\numstud \noisestd^2}{\numobs} \leq \minimax^{\cumlapnorm}_n(\textsc{\tstone}) \leq 
\frac{5}{\kappa^2} \frac{\numstud \noisestd^2}{\numobs}~.
\end{equation*}
\end{theorem}

\begin{theorem}[\btl]\label{thm:btl}
Assume that $\Lnorm{\mathbf{\wt^*}}{\infty} \leq \wmax$ (known) and $n \geq \frac{0.04467 \noisestd^2  \trace{\cumlapnorm^\dagger}}{\wmax^2}$. The minimax rate is bounded as
\begin{equation*}
0.001 \frac{\numstud \noisestd^2}{\numobs} \leq \minimax^{\cumlapnorm}_n(\textsc{\btl}) \leq 
1.37\left(e^\frac{\wmax}{\noisestd}+e^\frac{-\wmax}{\noisestd}\right)^4 \frac{\numstud \noisestd^2}{\numobs}~.
\end{equation*}
\end{theorem}
\end{subtheorem}
%
The upper bound in each case is from an analysis of the maximum likelihood (ML) estimator. The ML estimator, in all three settings, is the solution to a convex-optimization problem (while this is clear for the \pair~and \btl~models, see for instance \cite{tsukida2011analyze} for a proof in the \tstone~case). \\
\textbf{Proof Sketch: } 
{\bf Lower bound: } The lower bounds are based on a combination of information-theoretic techniques
and carefully constructed packings of the parameter set $\mathcal{W}$. Such techniques are standard 
in minimax analysis \cite{tsybakovbook}. The main technical difficulty is in constructing a packing in 
the semi-norm induced by $\cumlapnorm.$
A consequence of Fano's inequality (see
for instance
Theorem 2.5 in \cite{tsybakovbook}) is that
if we can construct a packing of vectors $\{\mathbf{w}_1,\ldots,\mathbf{w}_M\}$ of vectors 
in $\mathcal{W}$ such that (a) the KL divergence between the induced distributions is small, i.e. 
$\max_{ij} \kl{\mathbb{P}_{\mathbf{w}_i}}{\mathbb{P}_{\mathbf{w}_j}} \leq \beta \log M$
for a sufficient small (universal) constant $\beta$, and (b) $\min_{ij} \mnorm{\mathbf{w}_i - \mathbf{w}_j}{\cumlapnorm}^2 \geq \delta$ for some parameter $\delta$, then  for a small constant $c$, the minimax risk above is at least $c \delta$. The main effort 
is in constructing an \emph{exponentially} large (in $\numstud$) packing in the $\cumlapnorm$ norm
with sufficiently large $\delta$,
and bounding the model specific constants $\beta$ and $c$ above. 
The condition on $n$ is used to ensure that the constituents of the packing satisfy $\Lnorm{\mathbf{\wt}}{\infty}\leq \wmax$. We relegate the details 
to the Appendix.

{\bf Upper bound: } In each case we analyze the maximum likelihood estimator $\wtest = \amin_{ \mathbf{1}^T \mathbf{w} = 0} \ell(\mathbf{w})$ where $\ell: \mathcal{W} \mapsto \mathbb{R}$ is 
the negative log-likelihood under the corresponding model. In the case of the \btl~and \tstone~models we impose
the additional constraint that $\|\mathbf{w}\|_\infty \leq B.$

The optimization problem in each case is \emph{convex}. 
The analysis follows along the lines of standard statistical analyses of M-estimators \cite{svd}. We proceed by
upper and lower bounding
the quantity
$$f(\wtest) = \ell(\wtest) - \ell(\wtopt) - \inprod{\nabla \ell(\wtopt)}{ \wtest - \wtopt}$$
where $\nabla \ell(\wtopt) \in \mathbb{R}^d$ is the gradient of the negative log-likelihood.

In particular, an analysis of the \emph{strong convexity} parameter of the negative log-likelihood provides
a lower bound of the form
$f(\wtest) \geq \gamma \mnorm{\wtest - \wtopt}{\cumlapnorm}^2$
for an appropriate $\gamma$. Since $\wtest$ is the maximum-likelihood
estimator, we get $\ell(\wtest) \leq \ell(\wtopt)$. This implies
$f(\wtest) \leq - \inprod{\nabla \ell(\wtopt)}{ \wtest - \wtopt} \leq \mnorm{ \wtest - \wtopt}{\cumlapnorm} 
\mnorm{\nabla \ell(\wtopt)}{\cumlapnorm^\dagger}$
via Cauchy-Schwarz under appropriate conditions (recall that $\Sigma$ only induces a semi-norm).
Putting these together we arrive at the bound,
$\gamma\mnorm{\wtest - \wtopt}{\cumlapnorm} \leq \mnorm{\nabla \ell(\wtopt)}{\cumlapnorm^\dagger}.$
The main model-specific effort is in analyzing the strong convexity parameter and bounding 
the $\Sigma^\dagger$-norm of $\nabla \ell(\wtopt).$ We defer the details to the Appendix.
\qed

A minimax analysis of the BTL model is also provided in Negahban et al.~\cite{negahban2014rank}.
Although their main focus is the analysis of a random walk based algorithm, they also provide an analysis
for the MLE for the case of uniformly randomly chosen $\diff_i$. Their information theoretic lower bound studies a related but different problem. Their analysis applies only to the specific sampling schemes considered and show a considerable gap between the MLE and the lower bound. Our analysis however eliminates this discrepancy and shows that MLE is in fact minimax (rate) optimal for $\minimax_n^{\cumlapnorm}$.

To conclude this section, let us develop some consequences of this theorem. Let us focus on upper bounds
in the ordinal setting, and consider estimation error in $\ell^2_2$. As in the theorem, we assume 
that the graph induced 
by the comparisons is connected. 
Now ignoring model specific constants we can see that
$$\minimax_n^2 \leq \frac{ d \sigma^2}{n \lambda_2(\cumlapnorm)}$$
where $\lambda_2(\cumlapnorm)$ is the second smallest eigenvalue of $\cumlapnorm.$
Recall that $\cumlapnorm$ is simply the standardized Laplacian of the comparison graph, and its second
eigenvalue is determined by the \emph{topology} of the chosen comparisons. To understand this we consider
three canonical examples, and in each case we assume that the comparison graph is fixed, $\numobs$ is large enough 
and that the samples are distributed evenly along the fixed graph. It is straightforward to extend this to the case
of \emph{randomly} chosen comparisons from a fixed graph using matrix concentration inequalities 
(see for instance \cite{mchammer}).
\begin{enumerate}[leftmargin=0cm,itemindent=0.5cm,labelwidth=\itemindent,labelsep=0cm,align=left]
\item Dumbell graph: This is the graph on $\numstud$ vertices, which consists of two cliques of $\numstud/2$ disjoint sets of vertices with a single edge between them. Suppose $\numobs \geq {\numstud/2 \choose 2} + 1$. Since the unweighted graph has $\lambda_2 = O(1)$ we get \mbox{$\lambda_2(\cumlapnorm) = \frac{O(1)}{{\numstud/2 \choose 2} + 1}$} and the $\ell^2_2$
error scales as 
$\minimax_n^2 \leq \frac{ {{\numstud/2 \choose 2}} d \sigma^2}{\numobs}.$
\item Complete graph: Suppose $\numobs \geq {\numstud \choose 2}$. It is easy to verify that since the unweighted complete graph has $\lambda_2 = \numstud$, we get \mbox{$\lambda_2(\cumlapnorm) = \frac{ \numstud }{ {\numstud \choose 2}}$} and the $\ell^2_2$
error scales as 
$\minimax_n^2 \leq \frac{ {\numstud \choose 2} \sigma^2}{\numobs}.$
\item Degree-$k$ expander: The unweighted degree-$k$ expander has $\lambda_2 = O(k)$ and a similar argument as before shows that if $\numobs \geq k \numstud$ then we get the error scales as
\mbox{$\minimax_n^2 \leq \frac{ d^2 \sigma^2}{\numobs}.$}
\end{enumerate}
To summarize we see the $\ell_2^2$ error scaling of $\frac{d^2 \sigma^2}{\numobs}$ for the complete graph and 
the degree-$k$ expander. We conjecture that this is in fact the {best possible} scaling.
Observe that the degree-$k$ expander requires $n \geq k \numstud$ while the complete graph requires $\numobs \geq {d \choose 2}$, so in practical applications at least for small sample sizes, we should prefer a low-degree expander. On the other hand, for the dumbell graph, the 
error scales as $d^3 \sigma^2/n$ indicating that is a {bad} topology.

\section{Inference in the Experimental Data}\label{sec:inference}
In this section we return to our experimental data from Section~\ref{sec:experiments}. 
We consider data from the three experiments of identifying number of spelling errors, estimating the distances between cities, and recognizing the frequencies of audio, for which we know the ground truth. For each of the three experiments, we execute $100$ iterations of the following procedure. Select five workers from the cardinal and five from the ordinal pool  of workers who did this experiment, uniformly at random without replacement. (The number five is inspired by practical systems~\cite{wang2011managing,piech2013tuned}.) Run the maximum-likelihood estimator of the \dir~model on the data from the five workers selected from the cardinal pool, and the maximum-likelihood estimator of the \tstone~model on the data from the five workers of the ordinal pool. In particular, the estimator for the ordinal case first estimates $\noisestd$ via 3-fold cross-validation, choosing the value that maximizes held-out data log likelihood, and then uses this best fit for the rest of the estimation procedure. Note that unlike Section~\ref{sec:experiments}, the cardinal data here is \textit{not} converted to ordinal.

We evaluated the performance of these two estimators as follows. The true and inferred vectors were first scaled to have their maximum elements equal to $1$ and minimum elements equal to $-1$; this mimics the effect of knowing the scaling $\wmax$ via `domain knowledge'. The (scaled) inferred vectors in either case were then compared with the (scaled) true vector in terms of two metrics: (i) $\frac{1}{\numstud}$ times the squared $\ell_2$ distance, and (ii) the Kendall's tau rank correlation coefficient.

The results of this evaluation are enumerated in Table~\ref{tab:inference} (boldface indicates a better performance). To put the results in perspective of the rest of the paper, let us also recall the per-sample errors in these experiments from Table~\ref{tab:mturk_results}. Observe that in the experiment of estimating distances, the per-sample error in the cardinal data was significantly higher than the ordinal data. This is reflected in the results of Table~\ref{tab:inference} where the estimator on the ordinal data performs much better (in terms of the $\ell_2$ error) than the estimator on the cardinal data. On the other hand, the task of identifying the number of spelling mistakes involved a per-sample noise that was comparable across the two settings, and hence the estimator on the cardinal data scores over the ordinal one. As one would expect, the ordinal approach outperforms cardinal in terms of the (ordinal) Kendall's tau coefficient.




\begin{table}[h]
\begin{center}
\begin{tabular}{|l|c|c|c|}
\hline
Task & Spelling & Distance & Audio\\
\hline
Squared $\ell_2$-distance in Ordinal  & {0.358} & \textbf{0.168}  & \textbf{0.444}\\
Squared $\ell_2$-distance in Cardinal & \textbf{0.350 } & {0.330} &{ 0.508}\\
\hline
Kendall's tau coefficient in Ordinal & \textbf{0.277  }& \textbf{0.547} & \textbf{0.513}\\
Kendall's tau coefficient in Cardinal & 0.129  & 0.085 & 0.304\\
\hline
\end{tabular}
\end{center}
\caption{Evaluation of the inferred solution from the data received from multiple workers.}
\label{tab:inference}
\end{table}

\section{Conclusion}
This paper compares cardinal and ordinal approaches to evaluation performed by humans. With an increasing number of systems relying on non-expert human evaluators (e.g., using crowdsourcing), the choice of the evaluation mechanism forms a critical component of these systems. We argue by means of experiments and fundamental theoretical bounds that ordinal data provides a better estimate of the true solution when the per-sample noise is low enough relative to cardinal data, and the threshold for this choice is independent of the number of observations and the number of questions. 
This suggests a guideline for deciding whether to deploy a cardinal or an ordinal method of data collection: estimate the noise in the data by obtaining a few samples from either method, and then use the bounds on the overall error to determine the better of the two options.

We suggest further research to understand the tradeoffs in cardinal and ordinal measurements.
Our theoretical results were based on simple models, but more complex models, such as ones incorporating the abilities of the different human workers, could be more accurate.
Other model classes might have different noise thresholds determining when cardinal or ordinal performs best.
Also, it would be useful to make in-depth studies of noise in specific crowdsourcing settings, such as user experience testing and peer grading in classes.

Future research could also improve data collection.
For both cardinal and ordinal data, it would be useful to derive methods for adaptively choosing which measurements to take.
Our results on topology-aware bounds could potentially be used to improve ordinal evaluation by analyzing the best topologies for choosing pairs of items to compare.


\putbib
\end{bibunit}

\begin{bibunit}
\clearpage
\appendix
Appendix~\ref{app:mturk} provides some additional details on the experiments. Appendix~\ref{app:refresher} reviews some technical results that are used in our theoretical proofs. Appendix~\ref{app:proofs} presents proofs of the theoretical results.
\section{Additional Details on Experiments}\label{app:mturk}
This section presents additional details on the experiments presented in Section~\ref{sec:experiments} and Section~\ref{sec:inference}.  We first discuss the experiments of Section~\ref{sec:experiments}. The data was collected by putting up tasks on Amazon Mechanical Turk (mturk.com). Amazon Mechanical Turk is an online platform for putting up tasks, where any individual or institution can put up tasks and offer certain payments, and anyone can log in and complete the tasks in exchange for some payment that was specified along with the task. The following are some additional specifics about the experiments described in Section~\ref{sec:experiments}.
\begin{itemize}
\item Each experiment comprised of 100 tasks, all comprising the same set of questions but organized in either a cardinal or ordinal format at random. 
\item A worker was offered $20$ cents for any task she completed. 
\item A worker was allowed to do no more than one task in an experiment. 
\item Workers were required to answer all the questions in a task. 
\item Only those workers who had $100$ or more approved works prior to this and also had at least $95\%$ approval rate were allowed. 
\item Workers from any country were allowed to participate, except for the task of estimating distances between cities where only workers from the USA were allowed since all the questions were about American cities.
\end{itemize}

We now move on to discuss the inference algorithms of Section~\ref{sec:inference}. The inference algorithms in this section operated on the data from three of the experiments. The four remaining experiments were unsuitable for this purpose: the experiments on rating the relevance of search results and rating taglines for a product had no ground truth; the task of identifying the area of circles had each question drawn independently at random from a Beta distribution, and hence no two workers answered the same questions; the comparison graph for the experiment on identifying age from pictures was not a connected graph. 

Table~\ref{tab:inference} in Section~\ref{sec:inference} presented the average errors across $100$ runs of the inference procedure; Table~\ref{tab:inference_stddev} here tabulates the associated standard deviation of the errors across the $100$ runs.




\begin{table}[h]
\begin{center}
\begin{tabular}{|l|c|c|c|}
\hline
Task & Spelling & Distance & Audio\\
\hline
Squared $\ell_2$-distance in Ordinal  & 0.122 & 0.070  & 0.302\\
Squared $\ell_2$-distance in Cardinal & 0.207  & 0.076 & 0.279\\
\hline
Kendall's tau coefficient in Ordinal & 0.244  & 0.113 & 0.217\\
Kendall's tau coefficient in Cardinal & 0.214  & 0.148 & 0.239\\
\hline
\end{tabular}
\end{center}
\caption{The standard deviation of the errors incurred in the $100$ runs of the inference procedure of Section~\ref{sec:inference}. The average of the errors is listed in Table~\ref{tab:inference}.}
\label{tab:inference_stddev}
\end{table}

\section{Review of some Technical Results}\label{app:refresher}
In this section we present some well known information-theoretic 
results that we use in our proofs. See for instance \cite{tsybakovbook} for proofs
of these claims. 
\subsection{Fano's inequality: Multiple hypothesis version}

\blem
\label{lem::fano}
Let $X$ 
be a random variable with distribution equal to one of $r+1$ possible distributions
$\mathbb{P}_1,\ldots,\mathbb{P}_{r+1}$. 
Furthermore, the Kullback-Leibler divergence between any pair of densities cannot be too large,
$$\kl{\mathbb{P}_i}{\mathbb{P}_j} \leq \beta~~\forall~~i \neq j.$$
Let $\psi(X)\in\{1,\ldots, r+1\}$ be an estimate of the index. Then
$$\sup_i \mathbb{P}_i(\psi(X)\not = i) \geq 1-\frac{\beta+\log 2}{\log r}.$$
\elem

\subsection{Fano's inequality: Two hypothesis version}
\blem
Let $X$ be a random variable with distribution either $\mathbb{P}_0$ or $\mathbb{P}_1$, and suppose that
the KL divergence between $f_0$ and $f_1$ be bounded as
$$\kl{\mathbb{P}_0}{\mathbb{P}_1} \leq \alpha < \infty.$$
Then for any $\psi(X)\in \{0,1\}$ we have 
$$\sup_{i \in \{0,1\}} \mathbb{P}_i(\psi(X)\not = i) \geq \frac{ 1 - \sqrt{\alpha/2}}{2}.$$
\elem

\subsection{Estimation error}
Let $\mathcal{P}$ be a family of distributions.
Consider a map $\theta: \mathcal{P} \mapsto \Omega$, 
and let $\rho$ be a semi-norm on $\Omega$. 
An estimator $\hat{\theta}$ is a measurable function $\hat{\theta}: \mathcal{P} \mapsto \Omega$. 
The following Lemma gives a lower bound on
the minimax error in estimating $\theta$ in the metric induced by $\rho$.
\blem
Let $X$ 
be a random variable with distribution equal to one of $r+1$ possible distributions
$\mathbb{P}_1,\ldots,\mathbb{P}_{r+1}$ such that
$$\kl{\mathbb{P}_i}{\mathbb{P}_j} \leq \beta~~\forall~~i \neq j.$$
Suppose that $$\min_{ij} \rho(\theta(\mathbb{P}_i), \theta(\mathbb{P}_j)) \geq \delta$$
then the minimax estimation error of any estimator is lower bounded as
$$\inf_{\hat{\theta}} \sup_{i \in \{1,\ldots,r+1\}} \mathbb{E} [\rho(\hat{\theta}, \theta(\mathbb{P}_i))] \geq 
\frac{\delta}{2} \left( 1-\frac{\beta+\log 2}{\log r} \right).$$
\elem

\section{Proofs}\label{app:proofs}
We first  introduce some notation which will be employed subsequently in the proofs.
Observe that $\cumlap$ is a positive semi-definite matrix (recall from Section~\ref{sec::bounds}). Let 
\beq \cumlap = U \Lambda U^T~. \nonumber \eeq
Let $\lambda_1>\ldots>\lambda_\numstud$ be the eigenvalues of $\cumlap$ and assume without loss of generality that $\forall i \in [\numstud]$, $\lambda_i$ is the $(i,i)^{\textrm{th}}$ entry of $\Lambda$.
Since the graph topologies are assumed to be connected, we have $\lambda_i \neq 0~\forall~i\in[\numstud-1]$ and $\lambda_\numstud=0$.
The Moore-Penrose pseudoinverse of $\cumlap$ is the $(\numstud \times \numstud)$ matrix $\cumlapinv$, and this satisfies
\beq
\cumlapinv := U \tilde{\Lambda} U^T \quad \textrm{where}~\tilde{\lambda}_i = \lambda_i^{-1} \mathbf{1}\{\lambda_i \neq 0\}.~\nonumber
\eeq 
Note that $\cumlapinv$ is also positive semidefinite, has a rank equal to $(\numstud-1)$, and $Tr(\cumlap \cumlapinv) = \numstud-1$. Furthermore, $\Lambda \tilde{\Lambda}=\tilde{\Lambda}^{\frac{1}{2}} \Lambda \tilde{\Lambda}^{\frac{1}{2}} = \sum_{i=1}^{\numstud-1} \mathbf{e}_i \mathbf{e}_i^T$. 

The standardized versions of $\cumlap$ and $\cumlapinv$ are $\cumlapnorm := \frac{1}{n} \cumlap$ and $\cumlapinvnorm := \numobs \cumlapinv.$ Note that $\cumlapinvnorm$ is the Moore-Penrose pseudoinverse of $\cumlapnorm$.

We first state two lemmas that we will use to prove our results. Lemma~\ref{lem:packing} is used to prove lower bounds and Lemma~\ref{lem:pseudocauchyschwarz} is used to prove upper bounds. The proofs of the two Lemmas are provided at the end of this section.

\begin{lemma}\label{lem:packing}
For any $\delta>0$, $\alpha \in (0,1)$, $\beta \in (0,1)$ with $\beta = \frac{\log 2 + \alpha \log \alpha - \alpha}{2}$, there exist a set of $\packsize$ vectors $\{\mathbf{\wt}_1,\ldots,\mathbf{\wt}_{\packsize}\}$, each of length $\numstud$, such that every pair of vectors satisfies
\bean
\packdmin \delta^2 \leq (\mathbf{\wt}_i-\mathbf{\wt}_j)^T M (\mathbf{\wt}_i-\mathbf{\wt}_j) \leq 4\delta^2
\eean
and every vector in this set also satisfies
\bean
\mathbf{1}^T \mathbf{\wt}_i = 0~.
\eean
\end{lemma}

\begin{lemma}\label{lem:pseudocauchyschwarz}
Consider any $(\numstud \times \numstud)$ positive semidefinite matrix $\cumlap$, and any vectors $\mathbf{x}, \mathbf{y} \in \mathbb{R}^\numstud$ such that $\mathbf{x} \perp \textrm{nullspace}(\cumlap)$. If $\cumlapinv$ is the Moore-Penrose pseudoinverse of $\cumlap$, then
\bean
\mathbf{x}^T \mathbf{y} \leq \sqrt{\mathbf{x}^T\cumlap \mathbf{x}} \sqrt{\mathbf{y}^T\cumlapinv \mathbf{y}}~.
\eean
\end{lemma}

\textit{\bf Proof of Theorem~\ref{thm::cvo}}
In the cardinal case when each coordinate is measured the same number of times, the \dir~model reduces to the well-studied normal location model, for which the MLE is known to be the minimax estimator and its risk is straightforward to characterize (see \cite{tpe} for instance). 

In the ordinal case the result follows from Theorem~\ref{thm:tstone}, with $\cumlapnorm = \frac{2}{\numstud(\numstud-1)} \left(\numstud I	 - \mathbf{1}\mathbf{1}^T\right)$, i.e., an appropriately scaled Laplacian of the complete graph. We know that $\boldsymbol{1}^T\mathbf{\wt^*}=0$, and further observe in the proof of Theorem~\ref{thm:tstone} that it suffices to consider $\mathbf{\hat{\wt}}$ such that $\boldsymbol{1}^T\mathbf{\hat{\wt}}=0$. It follows that $\minimax^{\cumlapnorm}_n(\textsc{\tstone}) = \frac{2\numstud}{\numstud-1} \frac{\minimax^{2}_n(\textsc{\tstone})}{\numstud}$. The quantity $\minimax^{\cumlapnorm}_n(\textsc{\tstone})$ is bounded in Theorem~\ref{thm:tstone}.
\qed

\textit{\bf Proof of Theorem~\ref{thm:pair} (\pair):} \\
\textbf{Lower Bounds:} For any $\mathbf{\wt}_1$ and $\mathbf{\wt}_2$, the KL divergence between the distributions of $\mathbf{\obs}$ under $\mathbf{\wt}_1$ and $\mathbf{\wt}_2$ as the true values is
\bean
\kl{P_{\mathbf{\wt}_1}(\mathbf{\obs})}{P_{\mathbf{\wt}_2}(\mathbf{\obs} )} = \frac{1}{\noisestd^2} \Lnormsqr{\mathbf{\wt}_1}{\mathbf{\wt}_2}{\cumlap}~.
\eean

For any $\delta>0$, Lemma~\ref{lem:packing} constructs a packing $\{\mathbf{\wt}_1,\ldots,\mathbf{\wt}_{\packsize}\}$ such that every pair of distinct vectors $\mathbf{\wt}_i$ and $\mathbf{\wt}_j$ in this packing satisfies (with $\packdmin = \alphacard$ and $\beta = \betacard$)
\bean
\alphacard \delta^2 \leq (\mathbf{\wt}_i-\mathbf{\wt}_j)^T M (\mathbf{\wt}_i-\mathbf{\wt}_j) \leq 4\delta^2
\eean
and furthermore every vector in this set also satisfies
\bean
\mathbf{1}^T \mathbf{\wt}_i = 0~.
\eean

Given this packing, we have
\bean
\max_{i,j} \kl{P_{\mathbf{\wt}_i}(\mathbf{\obs})}{P_{\mathbf{\wt}_j}(\mathbf{\obs})} \leq \frac{4 \delta^2}{\noisestd^2}
\eean
and
\bean
\min_{i,j} \Lnormsqr{\mathbf{\wt}_i}{\mathbf{\wt}_j}{\cumlap} \geq \alphacard \delta^2~.
\eean
Using Fano's inequality, we get
\bean
\Lnormsqr{\mathbf{\hat{\wt}}}{\mathbf{\wt^*}}{\cumlap} &\geq& \frac{\alphacard}{2} \delta^2 \left(1 - \frac{\frac{4 \delta^2}{\noisestd^2}+\log 2}{\betacard \numstud}\right)
\eean
Choosing
\bean
\delta^2 = 0.0076\noisestd^2\numstud~,
\eean
bounding $\frac{\log 2}{\numstud} < 0.07$ whenever $\numstud > 9$, and noting that
\bean
\minimax^{\cumlapnorm}_n(\textsc{\tstone})  = \frac{1}{\numobs} \Lnormsqr{\mathbf{\hat{\wt}}}{\mathbf{\wt^*}}{\cumlap}~,
\eean
we get the desired result.

\textbf{Upper Bounds:} 
Define function $\loss$ as
\bean
\loss(\mathbf{\wt}) =  \sum_{i=1}^{\numobs} (\obs_i - \diff_i^T\mathbf{\wt})^2~.
\eean
Consider the maximum likelihood estimator
\begin{equation*}
\mathbf{\hat{\wt}} \in \amin_{\mathbf{\wt}} \loss (\mathbf{\wt})~.
\end{equation*}
The solution is not unique (since the objective is invariant to shifting of $\mathbf{\wt}$), and hence we impose an additional constraint
\bean
\mathbf{\hat{\wt}} \in \amin_{\mathbf{\wt}:\mathbf{\wt}^T\mathbf{1}=0} \loss (\mathbf{\wt})~.
\eean


This is a loss function for the maximum likelihood estimator (and needs to be minimized). Now, the gradient and Hessian of this loss function is
\bean
\nabla \loss (\mathbf{\wt}) = -2 \sum_{i=1}^{\numobs} (\obs_i - \diff_i^T\wt) \diff_i = -2 \diffmx^T \epsvec \\
\eean
\bean
\nabla^2 \loss (\mathbf{\wt}) = 2 \cumlap~.
\eean
The third and higher order derivatives of $\loss$ are zero.

Defining $\Delta := \hat{\mathbf{\wt}} - \mathbf{\wt^*}$, we have
\bean
\loss(\mathbf{\wt^*} + \boldsymbol{\Delta}) - \loss( \mathbf{\wt^*} ) - \inprod{\nabla \loss (\mathbf{\wt^*})}{\boldsymbol{\Delta}} &=& 2\boldsymbol{\Delta}^T \cumlap \boldsymbol{\Delta}~.
\eean
Also, since $\mathbf{\hat{\wt}}$ minimizes this loss function, we have
\bean
\loss(\mathbf{\wt^*} + \boldsymbol{\Delta}) - \loss( \mathbf{\wt^*} ) - \inprod{\nabla \loss (\mathbf{\wt^*})}{\boldsymbol{\Delta}} &\leq & - \inprod{\nabla \loss (\mathbf{\wt^*})}{\boldsymbol{\Delta}}\\
&\leq & \sqrt{\nabla \loss (\mathbf{\wt^*})^T \cumlapinv \nabla\loss (\mathbf{\wt^*})} \sqrt{ \boldsymbol{\Delta}^T \cumlap \boldsymbol{\Delta}}~
\eean
where the last equation follows from Lemma~\ref{lem:pseudocauchyschwarz} proved below.

We shall now upper bound the quantity $\nabla \loss (\mathbf{\wt^*})^T \cumlapinv \nabla\loss (\mathbf{\wt^*})$. We have
\bean
\nabla \loss (\mathbf{\wt^*})^T \cumlapinv \nabla\loss (\mathbf{\wt^*}) &=& \epsvec^T \diffmx \cumlapinv \diffmx^T \epsvec \\
&=& \norm{\tilde{\Lambda}^{\frac{1}{2}} U^T \diffmx^T \epsvec}^2~.
\eean
Now, $\tilde{\Lambda}^{\frac{1}{2}} U^T \diffmx^T \epsvec \sim N(0, \tilde{\Lambda}^{\frac{1}{2}} U \cumlap U^T \tilde{\Lambda}^{\frac{1}{2}})$ and hence
\bean
E[ \norm{\tilde{\Lambda}^{\frac{1}{2}} U^T \diffmx^T \epsvec}^2 ] &=& \trace{\tilde{\Lambda}^{\frac{1}{2}} U^T \cumlap U \tilde{\Lambda}^{\frac{1}{2}}}\\
&=& \numstud-1~.
\eean
We will use \cite[Proposition 1]{hsu2012tail} which says that for $\epsvec \sim N(0,\noisestd^2 I)$ and any matrix $A$,
\bea
P(\norm{A \epsvec}^2 / \noisestd^2 >  \trace{A^T A} + 2 \sqrt{\trace{(A^T A)^2}t} + 2\norm{A^T A}t ) \leq e^{-t}~\quad~\forall~t \geq 0 \label{eq:hsu_prop1}
\eea
In our setting, we have $A = \tilde{\Lambda}^{\frac{1}{2}} U^T \diffmx^T$ and hence 
\[\trace{A^TA} = \trace{\diffmx \cumlapinv \diffmx^T}=\trace{\cumlap \cumlapinv}=\numstud-1~,\]
\[\trace{(A^T A)^2} = \trace{\diffmx \cumlapinv \diffmx^T \diffmx \cumlapinv \diffmx^T} = \numstud-1~,\]
\[\opnorm{A}^2 = \amax_{\norm{\mathbf{v}}=1} \mathbf{v}^T \diffmx \cumlapinv \diffmx^T \mathbf{v} =  \amax_{ \norm{\mathbf{v}}=1} \mathbf{v}^T V \Sigma^T U^TU \tilde{\Lambda} U^T U \Sigma V \mathbf{v} = \amax_{ \norm{\mathbf{v}}=1} \mathbf{v}^T \Sigma^T  \tilde{\Lambda} \Sigma \mathbf{v} = 1~\]
where the last equation follows from setting the singular value decomposition of $\diffmx$ as $U \Sigma V^T$ and noting that by definition of $\cumlapinv$, we have $\Sigma^T \tilde{\Lambda} \Sigma = \sum_{i=1}^{\numstud-1} \mathbf{e}_i \mathbf{e}_i^T$.
Substituting these values, we have
\bean
P( \norm{\tilde{\Lambda}^{\frac{1}{2}} U^T \diffmx^T \epsvec}^2 / \noisestd^2 > (\numstud-1) + 2\sqrt{(\numstud-1)t} + 2t ) \leq e^{-t}~\quad~\forall~t \geq 0
\eean
\bea
\Rightarrow P( \norm{\tilde{\Lambda}^{\frac{1}{2}} U^T \diffmx^T \epsvec}^2 > 2 t \noisestd^2 \numstud ) \leq e^{-t}~\quad~\forall~t \geq 1 \label{eq:hsu_prop1_simplified}
\eea
Putting everything together, we get
\bean
\sqrt{\Delta^T \cumlap \Delta} \leq  \sqrt{\frac{\numstud \noisestd^2 t}{2}} \quad w.p. \geq 1 - e^{-t} ~\quad~\forall~t\geq 1~.
\eean
Squaring both sides and substituting $\cumlapnorm = \frac{1}{\numobs} \cumlap$, we get
In terms of the standardized Laplacian, we have
\bean
\Delta^T \cumlapnorm \Delta \leq  \frac{\numstud \noisestd^2 t}{2\numobs} \quad w.p. \geq 1 - e^{-t} ~\quad~\forall~t\geq 1~.
\eean
Finally,
\[
\frac{2\numobs}{\numstud \noisestd^2} \mathbb{E}\left[\Delta^T \cumlapnorm \Delta \right] = \int_{t=0}^{\infty} \mathbb{P}\left(\frac{2\numobs}{\numstud \noisestd^2} \Delta^T \cumlapnorm \Delta > t\right) \leq 1 + \int_{t=1}^{\infty} e^{-t} = 1 + \frac{1}{e}~.\]

\qed

\textit{\bf Proof of Theorem~\ref{thm:tstone} (\tstone):} \\
\textbf{Lower Bounds:} Let $\cdf$ denote the c.d.f. of the standard Gaussian distribution and let $\pdf$ denote its p.d.f. For any $\mathbf{\wt}_1$ and $\mathbf{\wt}_2$, the KL divergence between the distributions of $\mathbf{\obs}$ under $\mathbf{\wt}_1$ and $\mathbf{\wt}_2$ as the true values is
\bean
\kl{P_{\mathbf{\wt}_1}(\mathbf{\obs})}{ P_{\mathbf{\wt}_2}(\mathbf{\obs} ) }&=& \sum_{i=1}^{\numobs} \cdf(\mathbf{\wt}_1^T \diff_i/\noisestd) \log \frac{\cdf(\mathbf{\wt}_1^T \diff_i/\noisestd)}{\cdf(\mathbf{\wt}_2^T \diff_i/\noisestd)} + (1-\cdf(\mathbf{\wt}_1^T \diff_i/\noisestd)) \log \frac{1-\cdf(\mathbf{\wt}_1^T \diff_i/\noisestd)}{1-\cdf(\mathbf{\wt}_2^T \diff_i/\noisestd)}\nonumber\\
&\leq &\sum_{i=1}^{\numobs} (\cdf(\mathbf{\wt}_1^T \diff_i/\noisestd) - \cdf(\mathbf{\wt}_2^T \diff_i/\noisestd))  \frac{\cdf(\mathbf{\wt}_1^T \diff_i/\noisestd)}{\cdf(\mathbf{\wt}_2^T \diff_i/\noisestd)} \nonumber \\ 
&& + ((1-\cdf(\mathbf{\wt}_1^T \diff_i/\noisestd))-(1-\cdf(\mathbf{\wt}_2^T \diff_i/\noisestd)))  \frac{1-\cdf(\mathbf{\wt}_1^T \diff_i/\noisestd)}{1-\cdf(\mathbf{\wt}_2^T \diff_i/\noisestd)}\nonumber\\
&=& \sum_{i=1}^{\numobs}  \frac{(\cdf(\mathbf{\wt}_1^T\diff_i/\noisestd) - \cdf(\mathbf{\wt}_2^T\diff_i/\noisestd))^2}{\cdf(\mathbf{\wt}_2^T \diff_i/\noisestd) (1-\cdf(\mathbf{\wt}_2^T\diff_i/\noisestd))} \\ 
&\leq & \sum_{i=1}^{\numobs}  \frac{(\cdf(\mathbf{\wt}_1^T\diff_i/\noisestd) - \cdf(\mathbf{\wt}_2^T\diff_i/\noisestd))^2}{\cdf(2\wmax/\noisestd) (1-\cdf(2\wmax/\noisestd))} \\ 
&\leq &
\sum_{i=1}^{\numobs} \frac{f(0)^2}{\cdf(2\wmax/\noisestd) (1-\cdf(2\wmax/\noisestd))} {(\mathbf{\wt}_1^T\diff_i/\noisestd - \mathbf{\wt}_2^T\diff_i/\noisestd)^2}~.\\
&=&
\frac{1}{2\pi \noisestd^2 \cdf(2\wmax/\noisestd) (1-\cdf(2\wmax/\noisestd))} \Lnormsqr{\mathbf{\wt}_1}{\mathbf{\wt}_2}{\cumlap}~
\eean

For any $\delta>0$, Lemma~\ref{lem:packing} constructs a packing $\{\mathbf{\wt}_1,\ldots,\mathbf{\wt}_{\packsize}\}$ such that every pair of distinct vectors $\mathbf{\wt}_i$ and $\mathbf{\wt}_j$ in this packing satisfies (with $\packdmin = \alphacard$ and $\beta = \betacard$)
\bean
\alphacard \delta^2 \leq (\mathbf{\wt}_i-\mathbf{\wt}_j)^T M (\mathbf{\wt}_i-\mathbf{\wt}_j) \leq 4\delta^2
\eean
and furthermore every vector $\mathbf{\wt}_i$ in this set also satisfies
\bean
\mathbf{1}^T \mathbf{\wt}_i = 0~.
\eean

Given this packing, we have
\bean
\max_{i,j} \kl{P_{\mathbf{\wt}_i}(\mathbf{\obs})}{P_{\mathbf{\wt}_j}(\mathbf{\obs} )} \leq \frac{4 \delta^2}{2\pi \cdf(2\wmax/\noisestd) (1-\cdf(2\wmax/\noisestd))\noisestd^2}
\eean
and
\bean
\min_{i,j} \Lnormsqr{\mathbf{\wt}_i}{\mathbf{\wt}_j}{\cumlap} \geq \alphacard \delta^2~.
\eean
Using Fano's inequality, we get
\bean
\Lnormsqr{\mathbf{\hat{\wt}}}{\mathbf{\wt^*}}{\cumlap} &\geq& \frac{\alphacard}{2} \delta^2 \left(1 - \frac{\frac{4 \delta^2}{2\pi \cdf(2\wmax/\noisestd) (1-\cdf(2\wmax/\noisestd))\noisestd^2}+\log 2}{\betacard \numstud}\right)
\eean
Choosing
\bean
\delta^2 = 0.0076\noisestd^2\numstud \times 2\pi \cdf(2\wmax/\noisestd) (1-\cdf(2\wmax/\noisestd))
\eean
bounding $\frac{\log 2}{\numstud} < 0.07$ whenever $\numstud > 9$, and noting that
\bean
\minimax^{\cumlapnorm}_n(\textsc{\tstone})  = \frac{1}{\numobs} \Lnormsqr{\mathbf{\hat{\wt}}}{\mathbf{\wt^*}}{\cumlap}~,
\eean	
we get the desired result. The only issue remaining to consider is the bounded assumption of $\mathbf{\wt}$, and this is verified below.
\bea
 \|\mathbf{\wt}\|_\infty &=& \frac{\delta}{\sqrt{\numstud}} \|U \tilde{\Lambda}^\frac{1}{2} \mathbf{\wt}^{(2)}\|_\infty\nonumber\\
& \leq & \frac{\delta}{\sqrt{\numstud}} \sup_{\mathbf{u}:\norm{\mathbf{u}}=1} \mathbf{u}^T \tilde{\Lambda}^\frac{1}{2} \mathbf{\wt}^{(2)}\nonumber\\
&=& \frac{\delta}{\sqrt{\numstud}}  \frac{(\tilde{\Lambda}^\frac{1}{2} \mathbf{\wt}^{(2)})^T \tilde{\Lambda}^\frac{1}{2} \mathbf{\wt}^{(2)}}{\norm{\tilde{\Lambda}^\frac{1}{2} \mathbf{\wt}^{(2)}}}\nonumber\\
&\leq & \frac{\delta}{\sqrt{\numstud}}  \sqrt{\trace{\tilde{\Lambda}}}\label{eq:winbinary}\\
&=&\frac{\delta}{\sqrt{\numstud}}  \sqrt{\trace{\cumlapinv}}\nonumber\\
&=& \sqrt{0.00555\noisestd^2 \times 2\pi \cdf(2\wmax/\noisestd) (1-\cdf(2\wmax/\noisestd)) \frac{\trace{\cumlapinvnorm}}{\numobs}}\nonumber\\
&\leq& \wmax~,\nonumber
\eea
where~\eqref{eq:winbinary} follows from the fact that $\mathbf{\wt}^{(2)} \in \{-1,0,1\}^\numstud$ and the final equation follows from our assumption relating $n$ and $\trace{\cumlapinvnorm}$.

\textbf{Upper Bounds:} Define function $\loss$ as
\bean
\loss(\mathbf{\wt}) =  -\sum_{i=1}^{\numobs} \left[ \indicator{\obs_i=1} \log \cdf(\mathbf{\wt}^T \diff_i/\noisestd)+ \indicator{\obs_i=-1} \log (1-\cdf(\mathbf{\wt}^T \diff_i/\noisestd))\right]~.
\eean
Consider the maximum likelihood estimator
\bean
\mathbf{\hat{\wt}} \in \amin_{\mathbf{\wt}:\mathbf{1}^T \mathbf{\wt}=0, \|\mathbf{\wt}\|_\infty \leq \wmax} \loss (\mathbf{\wt})~.
\eean

The gradient and Hessian of this loss function are
\bean
\nabla \loss (\mathbf{\wt}) = \frac{-1}{\noisestd} \sum_{i=1}^{\numobs} \left[ \indicator{\obs_i=1} \frac{\pdf(\mathbf{\wt}^T \diff_i/\noisestd)}{\cdf(\mathbf{\wt}^T \diff_i/\noisestd)} - \indicator{\obs_i=-1} \frac{\pdf(\mathbf{\wt}^T \diff_i/\noisestd)}{1-\cdf(\mathbf{\wt}^T \diff_i/\noisestd)}\right]\diff_i~,
\eean
and
\bea
\nabla^2 \loss (\mathbf{\wt}) &=& \frac{1}{\noisestd^2} \sum_{i=1}^{\numobs} \left[ \indicator{\obs_i=1} \frac{\pdf(\mathbf{\wt}^T \diff_i/\noisestd)^2-\cdf(\mathbf{\wt}^T \diff_i/\noisestd) \pdf'(\mathbf{\wt}^T \diff_i/\noisestd)}{\cdf(\mathbf{\wt}^T \diff_i/\noisestd)^2} \right.\nonumber\\
& +& \left. \indicator{\obs_i=-1} \frac{\pdf(\mathbf{\wt}^T \diff_i/\noisestd)^2+(1-\cdf(\mathbf{\wt}^T \diff_i/\noisestd)) \pdf'(\mathbf{\wt}^T \diff_i/\noisestd)}{(1-\cdf(\mathbf{\wt}^T \diff_i/\noisestd))^2}\right] \diff_i \diff_i^T~\label{eq:hessian_comparative_threshold}
\eea
respectively. The scalar in the summation is always non-negative (since $\cdf$ is log-concave), and hence maximum likelihood inference is a convex optimization problem.

Define
\bean
\const_1(\noisestd,\wmax) &:=& \inf_{\mathbf{\wt}:\mathbf{1}^T\mathbf{\wt}=0,\|\mathbf{\wt}\|_\infty \leq \wmax, i \in [\numobs]} \min \left\{ \frac{\pdf(\mathbf{\wt}^T \diff_i/\noisestd)^2-\cdf(\mathbf{\wt}^T \diff_i/\noisestd) \pdf'(\mathbf{\wt}^T \diff_i/\noisestd)}{\cdf(\mathbf{\wt}^T \diff_i/\noisestd)^2}, \right. \nonumber \\
&&\left. \qquad\qquad\qquad\qquad\frac{\pdf(\mathbf{\wt}^T \diff_i/\noisestd)^2+(1-\cdf(\mathbf{\wt}^T \diff_i/\noisestd)) \pdf'(\mathbf{\wt}^T \diff_i/\noisestd)}{(1-\cdf(\mathbf{\wt}^T \diff_i/\noisestd))^2} \right\}\\
&=& \inf_{\mathbf{\wt}:\mathbf{1}^T\mathbf{\wt}=0,\|\mathbf{\wt}\|_\infty \leq \wmax, i \in [\numobs]} \frac{\pdf(\mathbf{\wt}^T \diff_i/\noisestd)^2+(1-\cdf(\mathbf{\wt}^T \diff_i/\noisestd)) \pdf'(\mathbf{\wt}^T \diff_i/\noisestd)}{(1-\cdf(\mathbf{\wt}^T \diff_i/\noisestd))^2}~\\
&=& \inf_{t \in [-2\wmax/\noisestd,2\wmax/\noisestd]}\frac{\pdf(t)^2+(1-\cdf(t)) \pdf'(t)}{(1-\cdf(t))^2}~\\
&=& \inf_{t \in [-2\wmax/\noisestd,2\wmax/\noisestd]} \left(\frac{\pdf(t)}{1-\cdf(t)}\right)^2-t\frac{\pdf(t)}{1-\cdf(t)}\\
&\geq& \left(\frac{t+\sqrt{t^2+\frac{8}{\pi}}}{2}\right)^2 - t\left(\frac{t+\sqrt{t^2+4}}{2}\right)\\
& = & \frac{2}{\pi} - \frac{t}{2}\left( \sqrt{t^2+4} - \sqrt{t^2 + \frac{8}{\pi}}\right)\\
& = & \frac{2}{\pi} - \frac{t}{2} \frac{ (t^2+4) - (t^2 + \frac{8}{\pi}) }{ \sqrt{t^2+4} + \sqrt{t^2 + \frac{8}{\pi}}}\\
& = & \frac{2}{\pi} - \left(2-\frac{4}{\pi}\right) \frac{ t }{ \sqrt{t^2+4} + \sqrt{t^2 + \frac{8}{\pi}}}\\
&\geq & \frac{4}{\pi}-1~.
\eean
Then for all $\mathbf{\wt}$ in the allowed set and any vector $\mathbf{v} \in \mathbb{R}^\numstud$, we have
\bean
\mathbf{v}^T \nabla^2 \loss (\mathbf{\wt}) \mathbf{v} &\geq & \frac{\const_1(\noisestd,\wmax)}{\noisestd^2} \sum_{i=1}^{\numobs}\mathbf{v}^T  \diff_i \diff_i^T~\mathbf{v}.
\eean

Defining $\Delta := \hat{\mathbf{\wt}} - \mathbf{\wt^*}$, we have
\bean
\loss(\mathbf{\wt^*} + \boldsymbol{\Delta}) - \loss( \mathbf{\wt^*} ) - \inprod{\nabla \loss (\mathbf{\wt^*})}{ \boldsymbol{\Delta}} & \geq & \boldsymbol{\Delta}^T \left(\frac{\const_1(\noisestd,\wmax)}{\noisestd^2} \sum_{i=1}^{\numobs} \diff_i \diff_i^T~.\right) \boldsymbol{\Delta}~\\
& = & \frac{\const_1(\noisestd,\wmax)}{\noisestd^2} \boldsymbol{\Delta}^T \cumlap \boldsymbol{\Delta}~.
\eean

Also, since $\mathbf{\hat{\wt}}$ minimizes this loss function, we have
\bean
\loss(\mathbf{\wt^*} + \boldsymbol{\Delta}) - \loss( \mathbf{\wt^*} ) - \inprod{\nabla \loss (\mathbf{\wt^*})}{\boldsymbol{\Delta}} &\leq & - \inprod{\nabla \loss (\mathbf{\wt^*})}{ \boldsymbol{\Delta}}\\
&\leq & \sqrt{\nabla \loss (\mathbf{\wt^*})^T \cumlapinv \nabla\loss (\mathbf{\wt^*})} \sqrt{ \boldsymbol{\Delta}^T \cumlap \boldsymbol{\Delta}}~
\eean
where the last equation follows from Lemma~\ref{lem:pseudocauchyschwarz}.

We will now upper bound the quantity $\nabla \loss (\mathbf{\wt^*})^T \cumlapinv \nabla\loss (\mathbf{\wt^*})$. Define independent random variables $\{\theta_i\}_{i=1}^{\numobs}$ as 
\bean
\theta_i = 
\begin{cases}
\frac{\pdf(\mathbf{\wt}^T \diff_i/\noisestd)}{\cdf(\mathbf{\wt}^T \diff_i/\noisestd)} &\textrm{w.p. \quad}\cdf(\mathbf{\wt}^T \diff_i/\noisestd)\\
\frac{-\pdf(\mathbf{\wt}^T \diff_i/\noisestd)}{1-\cdf(\mathbf{\wt}^T \diff_i/\noisestd)} &\textrm{w.p. \quad}1-\cdf(\mathbf{\wt}^T \diff_i/\noisestd)
\end{cases}~
\eean
and let $\boldsymbol{\theta}^T = [\theta_1,\cdots,\theta_\numobs]$.
Then
\bean
\nabla \loss(\mathbf{\wt}) = \frac{-1}{\noisestd} \diffmx^T \boldsymbol{\theta}~,
\eean
and
\bean
\nabla \loss (\mathbf{\wt^*})^T \cumlapinv \nabla\loss (\mathbf{\wt^*}) &=& \frac{1}{\noisestd^2} \boldsymbol{\theta}^T \diffmx \cumlapinv \diffmx^T \boldsymbol{\theta} \\
&=& \frac{1}{\noisestd^2} \norm{\tilde{\Lambda}^{\frac{1}{2}} U^T \diffmx^T \boldsymbol{\theta}}^2~.
\eean

We will now apply~\cite[Theorem 2.1]{hsu2012tail} which says that any random vector $\boldsymbol{\eps}$ that is zero-mean and sub-gaussian with parameter $\noisestd$, and any matrix $A$, must satisfy~\eqref{eq:hsu_prop1}. We will now set $\boldsymbol{\theta}$ as $\boldsymbol{\eps}$ and $\tilde{\Lambda}^{\frac{1}{2}} U^T \diffmx^T$ as $A$ in~\eqref{eq:hsu_prop1}. To this end, we see that 
\bean
\mathbf{E}\left[\boldsymbol{\theta}\right] = \mathbf{0}
\eean
and by virtue of each coordinate being being bounded, $\boldsymbol{\theta}$ is sub-gaussian with parameter at most $\const_2(\wmax,\noisestd)$ where $\const_2(\wmax,\noisestd)$ is defined as
\bean
\const_2(\wmax,\noisestd) &=& \sup_{\mathbf{\wt}:\mathbf{1}^T\mathbf{\wt}=0,\|\mathbf{\wt}\|_\infty \leq \wmax} \max_{i \in [\numobs]} \frac{ \pdf(\mathbf{\wt}^T \diff_i/\noisestd)}{\cdf(\mathbf{\wt}^T \diff_i/\noisestd)(1-\cdf(\mathbf{\wt}^T \diff_i/\noisestd))}~\\
&\leq& \frac{1}{\sqrt{2\pi} \cdf(2\wmax/\noisestd)(1-\cdf(2\wmax/\noisestd))}~.
\eean
Substituting these in~\eqref{eq:hsu_prop1} and following the simplifications of~\eqref{eq:hsu_prop1_simplified}, we get
\bean
P\left( \norm{\tilde{\Lambda}^{\frac{1}{2}} U^T \diffmx^T \epsvec}^2 > 2 t \const_2(\wmax,\noisestd)^2 \numstud \right) \leq e^{-t}~\quad~\forall~t \geq 1~.
\eean

Putting everything together, we have 
\bean
\boldsymbol{\Delta}^T \cumlap \boldsymbol{\Delta} \leq \frac{\const_2(\wmax,\noisestd)^2}{\const_1(\wmax,\noisestd)^2}2 \numstud \noisestd^2t \quad w.p. \geq 1-e^{-t} \quad \forall t \geq 1~.
\eean
Substituting the bounds on $\const_1$ and $\const_2$, and substituting $\cumlapnorm = \frac{1}{\numobs} \cumlap$, we get
\bean
\boldsymbol{\Delta}^T \cumlapnorm \boldsymbol{\Delta} \leq \frac{3.66}{\left(\cdf(2\wmax/\noisestd)(1-\cdf(2\wmax/\noisestd))\right)^2} \frac{\numstud \noisestd^2t}{\numobs} \quad w.p. \geq 1-e^{-t} \quad \forall t \geq 1~.
\eean

Converting this to a bound on $\mathbb{E}\left[\boldsymbol{\Delta}^T \cumlapnorm \boldsymbol{\Delta} \right]$ as done in the final step of the proof of Theorem~\ref{thm:pair} gives the desired result.
\qed

\textit{\bf Proof of Theorem~\ref{thm:btl} (\btl):}
\newcommand{\diffindexplus}{a}
\newcommand{\diffindexminus}{b}
\newcommand{\glmresidual}{\Psi}
For any differencing vector $\diff$, let $\diffindexplus(\diff)$ be the index of the `1' in $\diff$ and let $\diffindexminus(\diff)$ be the index of the `-1' in $\diff$. Now define a function $\glmresidual:\reals^\numstud \times \{-1,0,1\}^\numstud \rightarrow \reals$, where the second argument is always a differencing vector, as
\beq
\glmresidual(\mathbf{\wt},\diff) = \log\left(\exp\left(\frac{\wt_{\diffindexplus(\diff)}}{\noisestd}\right) + \exp\left(\frac{\wt_{\diffindexminus(\diff)}}{\noisestd}\right) \right) - \frac{\wt_{\diffindexplus(\diff)} + \wt_{\diffindexminus(\diff)}}{2\noisestd}~.\nonumber
\eeq

First, consider a single sample with observation $\obs$ and differencing vector $\diff$. We can rewrite the likelihood function of the BTL model as
\beq
P(\obs|  \wt) = \exp{\left( \frac{\obs}{2\noisestd} (\mathbf{\wt})^T\diff -\glmresidual(\mathbf{\wt},\diff) \right)}~.\nonumber
\eeq
Using this form, one can compute that
\begin{multline}
D_{KL}(P_{\mathbf{\wt}_1}({\obs}) || P_{\mathbf{\wt}_2}({\obs} ) ) = \frac{1}{2\noisestd} \frac{1-e^\frac{(\mathbf{\wt}_1)^T\diff}{\noisestd}}{1+e^\frac{(\mathbf{\wt}_1)^T\diff}{\noisestd}} (\mathbf{\wt}_1-\mathbf{\wt}_2)^T \diff \\
- \left((\mathbf{\wt}_1-\mathbf{\wt}_2)^T \nabla\glmresidual(\mathbf{\wt_1},\diff) + (\mathbf{\wt}_1-\mathbf{\wt}_2)^T \nabla^2\glmresidual(\mathbf{\wt_3},\diff)(\mathbf{\wt}_1-\mathbf{\wt}_2)\right)~,
\end{multline}
for some $\mathbf{\wt}_3$. One can evaluate that
\beq
\nabla\glmresidual(\mathbf{\wt_1},\diff) = \frac{1}{2\noisestd} \frac{1-e^\frac{(\mathbf{\wt}_1)^T\diff}{\noisestd}}{1+e^\frac{(\mathbf{\wt}_1)^T\diff}{\noisestd}}\diff \nonumber
\eeq
and that
\bea
\nabla^2 \glmresidual(\mathbf{\wt_3} ,\diff) &=& \frac{1}{2\noisestd^2}  \frac{1}{e^\frac{(\mathbf{\wt}_3)^T\diff}{\noisestd}+e^\frac{-(\mathbf{\wt}_3)^T\diff}{\noisestd}+2}\diff \diff^T\nonumber\\
& \leq & \frac{1}{8\noisestd^2} \diff \diff^T~.\nonumber
\eea
It follows that 
\beq
D_{KL}(P_{\mathbf{\wt}_1}({\obs}) || P_{\mathbf{\wt}_2}({\obs} ) ) \leq \frac{1}{8\noisestd^2} (\mathbf{\wt}_1-\mathbf{\wt}_2)^T \diff \diff^T (\mathbf{\wt}_1-\mathbf{\wt}_2)~.\nonumber
\eeq
Aggregating this over all samples, and observing that the distribution of the observation is independent across samples, we get
\beq
D_{KL}(P_{\mathbf{\wt}_1}({\obs}) || P_{\mathbf{\wt}_2}({\obs} ) ) \leq \frac{1}{8\noisestd^2} (\mathbf{\wt}_1-\mathbf{\wt}_2)^T \cumlap (\mathbf{\wt}_1-\mathbf{\wt}_2)~.\nonumber
\eeq

For any $\delta>0$, Lemma~\ref{lem:packing} constructs a packing $\{\mathbf{\wt}_1,\ldots,\mathbf{\wt}_{\packsize}\}$ such that every pair of distinct vectors $\mathbf{\wt}_i$ and $\mathbf{\wt}_j$ in this packing satisfies (with $\packdmin = \alphacard$ and $\beta = \betacard$)
\[
\alphacard \delta^2 \leq (\mathbf{\wt}_i-\mathbf{\wt}_j)^T M (\mathbf{\wt}_i-\mathbf{\wt}_j) \leq 4\delta^2
\]
and furthermore every vector in this set also satisfies
\[
\mathbf{1}^T \mathbf{\wt}_i = 0~.
\]

Given this packing, we have 
\beq
\max_{i,j} D_{KL}(P_{\mathbf{\wt}_1}(\mathbf{\obs}) || P_{\mathbf{\wt}_2}(\mathbf{\obs} ) ) \leq \frac{\delta^2}{2\noisestd^2}\nonumber
\eeq
and
\beq
\min_{i,j} \Lnormsqr{\mathbf{\wt}_1}{\mathbf{\wt}_2}{\cumlap} \geq \alphacard \delta^2~.
\eeq
Using Fano's inequality, we get
\bea
\Lnormsqr{\mathbf{\hat{\wt}}}{\mathbf{\wt^*}}{\cumlap} &\geq & \frac{\alphacard}{2} \delta^2 \left(1 - \frac{\frac{\delta^2}{2\noisestd^2}+\log 2}{\betacard \numstud}\right)
\eea
Choosing
\beq
\delta^2 = 0.06\noisestd^2\numstud~,
\eeq
bounding $\frac{\log 2}{\numstud} < 0.07$ whenever $\numstud > 9$, and noting that
\bean
\minimax^{\cumlapnorm}_n(\textsc{\tstone})  = \frac{1}{\numobs} \Lnormsqr{\mathbf{\hat{\wt}}}{\mathbf{\wt^*}}{\cumlap}~,
\eean
we get the desired result. The only issue remaining to consider is the boundedness of $\mathbf{\wt}$, and this is verified below.
\bea
\|\mathbf{\wt}\|_\infty &=& \frac{\delta}{\sqrt{\numstud}} \|U \tilde{\Lambda}^\frac{1}{2} \mathbf{\wt}^{(2)}\|_\infty\nonumber\\
& \leq & \frac{\delta}{\sqrt{\numstud}} \sup_{\mathbf{u}:\norm{\mathbf{u}}=1} \mathbf{u}^T \tilde{\Lambda}^\frac{1}{2} \mathbf{\wt}^{(2)}\nonumber\\
&=& \frac{\delta}{\sqrt{\numstud}}  \frac{(\tilde{\Lambda}^\frac{1}{2} \mathbf{\wt}^{(2)})^T \tilde{\Lambda}^\frac{1}{2} \mathbf{\wt}^{(2)}}{\norm{\tilde{\Lambda}^\frac{1}{2} \mathbf{\wt}^{(2)}}}\nonumber\\
&\leq & \frac{\delta}{\sqrt{\numstud}}  \sqrt{\trace{\tilde{\Lambda}}}\label{eq:winbinary_BTL}\\
&=&\frac{\delta}{\sqrt{\numstud}}  \sqrt{\trace{\cumlapinv}}\nonumber\\
&=& \sqrt{0.04467\noisestd^2  \frac{\trace{\cumlapinvnorm}}{\numobs}}\nonumber\\
&\leq& \wmax~, \nonumber
\eea
where~\eqref{eq:winbinary_BTL} follows from the fact that $\mathbf{\wt}^{(2)} \in \{-1,0,1\}^\numstud$ and the final equation follows from our assumption relating $n$ and $\trace{\cumlapinvnorm}$.

\textbf{Upper Bounds:}
Define function $\loss$ as
\beq
\loss(\mathbf{\wt}) =  \sum_{i=1}^{\numobs} \log\left(1+\exp\left(\frac{-\obs_i \mathbf{\wt}^T\diff_i}{\noisestd}\right)\right)~.\nonumber
\eeq
Consider the maximum likelihood estimator
\beq
\mathbf{\hat{\wt}} \in \amin_{\mathbf{\wt}:\mathbf{1}^T \mathbf{\wt}=0, \|\mathbf{\wt}\|_\infty \leq \wmax} \loss (\mathbf{\wt})~.\nonumber
\eeq

The gradient and Hessian of this loss function are
\beq
\nabla \loss (\mathbf{\wt}) = \frac{1}{\noisestd} \sum_{i=1}^{\numobs} \frac{-\obs_i e^{\frac{ -\obs_i \mathbf{\wt}^T\diff_i}{\noisestd}}}{1+e^{\frac{-\obs_i \mathbf{\wt}^T\diff_i}{\noisestd}}}\diff_i\nonumber
\eeq
\bea
\nabla^2 \loss (\mathbf{\wt}) &=& \frac{1}{\noisestd^2} \sum_{i=1}^{\numobs} \frac{e^{\frac{-\obs_i \mathbf{\wt}^T\diff_i}{\noisestd}}}{\left(1+e^{\frac{-\obs_i \mathbf{\wt}^T\diff_i}{\noisestd}}\right)^2} \diff_i \diff_i^T~.\nonumber
\eea
One can see that the Hessian is positive semi-definite, making function $\loss$ a convex function.

Then for all $\mathbf{\wt}$ in the allowed set, any observation $\obs_i \in \{-1,1\}$ and any differencing vector $\diff_i$, it must be that
\beq
\frac{e^{\frac{-\obs_i \mathbf{\wt}^T\diff_i}{\noisestd}}}{\left(1+e^{\frac{-\obs_i \mathbf{\wt}^T\diff_i}{\noisestd}}\right)^2} \geq \frac{1}{\left(e^\frac{\wmax}{\noisestd}+e^\frac{-\wmax}{\noisestd}\right)^2}~.\nonumber
\eeq

Defining $\Delta := \hat{\mathbf{\wt}} - \mathbf{\wt^*}$, we have
\bea
\loss(\mathbf{\wt^*} + \boldsymbol{\Delta}) - \loss( \mathbf{\wt^*} ) - \inprod{\nabla \loss (\mathbf{\wt^*})}{ \boldsymbol{\Delta}} & \geq &\boldsymbol{\Delta}^T \nabla^2 \loss (\mathbf{\wt}) \boldsymbol{\Delta}\nonumber\\
&\geq & \boldsymbol{\Delta}^T \left(\frac{1}{\noisestd^2 \left(e^\frac{\wmax}{\noisestd}+e^\frac{-\wmax}{\noisestd}\right)^2} \sum_{i=1}^{\numobs} \diff_i \diff_i^T~.\right) \boldsymbol{\Delta}~\nonumber\\
& = & \frac{1}{\noisestd^2 \left(e^\frac{\wmax}{\noisestd}+e^\frac{-\wmax}{\noisestd}\right)^2} \boldsymbol{\Delta}^T \cumlap \boldsymbol{\Delta}~.\nonumber
\eea

Also, since $\mathbf{\hat{\wt}}$ minimizes this loss function, we have
\bea
\loss(\mathbf{\wt^*} + \boldsymbol{\Delta}) - \loss( \mathbf{\wt^*} ) - \inprod{\nabla \loss (\mathbf{\wt^*})}{\boldsymbol{\Delta}} &\leq & - \inprod{\nabla \loss (\mathbf{\wt^*})}{ \boldsymbol{\Delta}}\nonumber\\
&\leq & \sqrt{\nabla \loss (\mathbf{\wt^*})^T \cumlapinv \nabla\loss (\mathbf{\wt^*})} \sqrt{ \boldsymbol{\Delta}^T \cumlap \boldsymbol{\Delta}}~\nonumber
\eea
where the last equation follows from Lemma~\ref{lem:pseudocauchyschwarz}.

We will now upper bound the quantity $\nabla \loss (\mathbf{\wt^*})^T \cumlapinv \nabla\loss (\mathbf{\wt^*})$. Define independent random variables $\{\theta_i\}_{i=1}^{\numobs}$ as 
\beq
\theta_i = 
\begin{cases}
\frac{-e^{\frac{\mathbf{\wt}^T\diff_i}{\noisestd}}}{1+e^{\frac{\mathbf{\wt}^T\diff_i}{\noisestd}}} &\textrm{w.p. \quad}\frac{1}{1+e^{\frac{\mathbf{\wt}^T\diff_i}{\noisestd}}}\\
~\vspace{-9pt}\\
\frac{e^{\frac{-\mathbf{\wt}^T\diff_i}{\noisestd}}}{1+e^{\frac{-\mathbf{\wt}^T\diff_i}{\noisestd}}} &\textrm{w.p. \quad}\frac{1}{1+e^{\frac{-\mathbf{\wt}^T\diff_i}{\noisestd}}}
\end{cases}~
\nonumber
\eeq
and let $\boldsymbol{\theta}^T = [\theta_1,\cdots,\theta_\numobs]$.
Then
\beq
\nabla \loss(\mathbf{\wt}) = \frac{-1}{\noisestd} \diffmx^T \boldsymbol{\theta}~,\nonumber
\eeq
and
\bea
\nabla \loss (\mathbf{\wt^*})^T \cumlapinv \nabla\loss (\mathbf{\wt^*}) &=& \frac{1}{\noisestd^2} \boldsymbol{\theta}^T \diffmx \cumlapinv \diffmx^T \boldsymbol{\theta} \\
&=& \frac{1}{\noisestd^2} \norm{\tilde{\Lambda}^{\frac{1}{2}} U^T \diffmx^T \boldsymbol{\theta}}^2~.\nonumber
\eea

We will now apply~\cite[Theorem 2.1]{hsu2012tail} which says that any random vector $\boldsymbol{\eps}$ that is zero-mean and sub-gaussian with parameter $\noisestd$, and any matrix $A$, must satisfy~\eqref{eq:hsu_prop1}. We will now  set $\boldsymbol{\theta}$ as $\boldsymbol{\eps}$ and $\tilde{\Lambda}^{\frac{1}{2}} U^T \diffmx^T$ as $A$ in~\eqref{eq:hsu_prop1}. To this end, we see that 
\beq
\mathbf{E}\left[\boldsymbol{\theta}\right] = \mathbf{0}\nonumber
\eeq
and by virtue of each coordinate being being bounded, $\boldsymbol{\theta}$ is sub-gaussian. The sub-gaussianity parameter is upper bounded by
\[
\frac{e^{\frac{-\mathbf{\wt}^T\diff_i}{\noisestd}}}{1+e^{\frac{-\mathbf{\wt}^T\diff_i}{\noisestd}}}-\frac{-e^{\frac{\mathbf{\wt}^T\diff_i}{\noisestd}}}{1+e^{\frac{\mathbf{\wt}^T\diff_i}{\noisestd}}} = 1~.
\]

Substituting these in~\eqref{eq:hsu_prop1} and following the simplifications of~\eqref{eq:hsu_prop1_simplified}, we get
\beq
P( \norm{\tilde{\Lambda}^{\frac{1}{2}} U^T \diffmx^T \epsvec}^2 > 2 t  \numstud ) \leq e^{-t}~\quad~\forall~t \geq 1~.\nonumber
\eeq

Putting everything together, we have 
\beq
\sqrt{\boldsymbol{\Delta}^T \cumlap \boldsymbol{\Delta}} \leq  \left(e^\frac{\wmax}{\noisestd}+e^\frac{-\wmax}{\noisestd}\right)^2 \sqrt{2 \numstud \noisestd^2t} \quad w.p. \geq 1-e^{-t} \quad \forall t \geq 1~.\nonumber
\eeq

Squaring and substituting $\cumlapnorm = \frac{1}{\numobs} \cumlap$, we get
\beq
\boldsymbol{\Delta}^T \cumlapnorm \boldsymbol{\Delta} \leq \left(e^\frac{\wmax}{\noisestd}+e^\frac{-\wmax}{\noisestd}\right)^4 \frac{\numstud \noisestd^2 t}{\numobs} \quad w.p. \geq 1-e^{-t} \quad \forall t \geq 1~.\nonumber
\eeq
Converting this to a bound on $\mathbb{E}\left[\boldsymbol{\Delta}^T \cumlapnorm \boldsymbol{\Delta} \right]$ as done in the final step of the proof of Theorem~\ref{thm:pair} gives the desired result.
\qed

\textit{\bf Proof of Lemma~\ref{lem:packing}:} First construct a set of $\packsize$ vectors $\{\mathbf{\wt}_1^{(1)},\ldots,\mathbf{\wt}_{\packsize}^{(1)}\}$, each belonging to $\{-1,+1\}^{\numstud-1}$ such that
\bean
\min_{i \neq j} \textrm{Hamming-distance}(\mathbf{\wt}_i^{(1)}-\mathbf{\wt}_j^{(1)}) \geq \packdmin\numstud~.
\eean
The existence of such a set is guaranteed by the Gilbert-Varshamov bound, which guarantees existence of a (binary) code of length $(\numstud-1)$, minimum Hamming distance $\packdmin \numstud$, and the number of code words at least
\bean
\frac{2^{\numstud-1}}{\sum_{\ell=0}^{\packdmin\numstud-1}{\numstud-1 \choose \ell}} &\geq & 2^{\numstud-1}\left(\frac{\packdmin \numstud -1}{(\numstud-1)e}\right)^{\packdmin \numstud-1}\\
& = & e^{(\numstud-1)\log 2 + (\packdmin \numstud-1)\log \left(\frac{\packdmin \numstud-1}{e\numstud-e}\right)}\\
& \geq & e^{\frac{d}{2}\left(\log 2 + \packdmin \log \left(\frac{\packdmin}{e}\right)\right)}\\
&= & \packsize~.
\eean

It follows from the construction that for every pair of distinct vectors in this set,
\bean
4\packdmin \numstud \leq (\mathbf{\wt}_i^{(1)}-\mathbf{\wt}_j^{(1)})^T (\mathbf{\wt}_i^{(1)}-\mathbf{\wt}_j^{(1)}) \leq 4 \numstud~.
\eean

Now construct a second set of $\packsize$ vectors $\{\mathbf{\wt}_1^{(2)},\ldots,\mathbf{\wt}_{\packsize}^{(2)}\}$, each of length $\numstud$, as
\bean
\left(\mathbf{\wt}_i^{(2)}\right)^T = \left[ \left(\mathbf{\wt}_i^{(1)}\right)^T~~0\right]^T \quad \forall i ~.
\eean
It is easy to see that every pair of distinct vectors in this set satisfies
\bean
4\packdmin\numstud \leq (\mathbf{\wt}_i^{(2)}-\mathbf{\wt}_j^{(2)})^T (\mathbf{\wt}_i^{(2)}-\mathbf{\wt}_j^{(2)}) \leq 4\numstud~.
\eean
Finally, construct a third set of $\packsize$ vectors $\{\mathbf{\wt}_1,\ldots,\mathbf{\wt}_{\packsize}\}$, each of length $\numstud$, as
\bean
\mathbf{\wt}_i = \frac{\delta}{\sqrt{\numstud}} U \tilde{\Lambda}^\frac{1}{2} \mathbf{\wt}_i^{(2)} \quad \forall i ~.
\eean
For any vector in this set
\bean
\mathbf{1}^T \mathbf{\wt}_i &=& \frac{\delta}{\sqrt{\numstud}} \mathbf{1}^T U \tilde{\Lambda}^\frac{1}{2} \mathbf{\wt}_i^{(2)}\\
&=& \frac{\delta}{\sqrt{\numstud}} \mathbf{e}_\numstud \mathbf{e}_\numstud^T \tilde{\Lambda}^\frac{1}{2} \mathbf{\wt}_i^{(2)}\\
&=&0~.
\eean
For any pair of vectors in this set,
\bean
(\mathbf{\wt}_i-\mathbf{\wt}_j)^T \cumlap (\mathbf{\wt}_i-\mathbf{\wt}_j) &=& \frac{\delta^2}{\numstud} (\mathbf{\wt}_i^{(2)}-\mathbf{\wt}_j^{(2)})^T  \tilde{\Lambda}^\frac{1}{2} U^T \cumlap U \tilde{\Lambda}^\frac{1}{2} (\mathbf{\wt}_i^{(2)}-\mathbf{\wt}_j^{(2)})\\
&=& \frac{\delta^2}{\numstud} (\mathbf{\wt}_i^{(2)}-\mathbf{\wt}_j^{(2)})^T  \tilde{\Lambda}^\frac{1}{2} \Lambda \tilde{\Lambda}^\frac{1}{2} (\mathbf{\wt}_i^{(2)}-\mathbf{\wt}_j^{(2)})\\
&=& \frac{\delta^2}{\numstud} (\mathbf{\wt}_i^{(2)}-\mathbf{\wt}_j^{(2)})^T (\mathbf{\wt}_i^{(2)}-\mathbf{\wt}_j^{(2)})
\eean
where the last step makes use of the fact that the last coordinate of each vector in the set $\{\mathbf{\wt}_1^{(2)},\ldots,\mathbf{\wt}_{\packsize}^{(2)}\}$ is zero.
It follows that
\bean
4\packdmin \delta^2 \leq  (\mathbf{\wt}_i-\mathbf{\wt}_j)^T \cumlap (\mathbf{\wt}_i-\mathbf{\wt}_j)  \leq 4 \delta^2~.
\eean
\qed

\textit{\bf Proof of Lemma~\ref{lem:pseudocauchyschwarz}:} Consider the singular value decompositions $\cumlap = U \Lambda U^T$, $\cumlapinv = U \tilde{\Lambda} U^T$. Let $\tilde{\mathbf{x}} := \Lambda^{\frac{1}{2}} U^T \mathbf{x}$ and $\tilde{\mathbf{y}} := \tilde{\Lambda}^{\frac{1}{2}} U^T \mathbf{y}$. Then
\bean
\sqrt{\mathbf{x}^T \cumlap \mathbf{x}} \sqrt{\mathbf{y}^T \cumlapinv \mathbf{y}} &=& \sqrt{\mathbf{x}^T U \Lambda U^T \mathbf{x}} \sqrt{\mathbf{y}^T U \tilde{\Lambda} U^T \mathbf{y}}\\
&=& \norm{\tilde{\mathbf{x}}} \norm{\tilde{\mathbf{y}}} \\
&\geq & \tilde{\mathbf{x}}^T \tilde{\mathbf{y}}\\
&=& \mathbf{x}^T U \Lambda^\frac{1}{2}\tilde{\Lambda}^\frac{1}{2}U^T\mathbf{y}\\
&=& \mathbf{x}^T U U^T\mathbf{y} \quad \textrm{(since $\mathbf{x} \perp$ nullspace$(\cumlap)$)}\\
&=& \mathbf{x}^T \mathbf{y}~.
\eean
\qed

\putbib
\end{bibunit}

\end{document}